\pdfoutput=1

\documentclass[11pt]{article}
\usepackage[utf8]{inputenc}

\usepackage[final]{acl}
\usepackage{breakurl} 
\usepackage{times}
\usepackage{latexsym}
\usepackage{dsfont}
\usepackage{amsmath}
\usepackage[T1]{fontenc}

\usepackage[utf8]{inputenc}

\usepackage{microtype}

\usepackage{inconsolata}

\usepackage{graphicx}
\usepackage{float}
\usepackage{multirow}
\usepackage{adjustbox}
\usepackage{booktabs}
\usepackage{amssymb}
\usepackage{amsmath}
\usepackage{subcaption}
\usepackage{xcolor}
\usepackage{fancyvrb}
\usepackage{lipsum}
\usepackage{listings}
\usepackage{xcolor}
\newtheorem{theorem}{Theorem}

\definecolor{green}{RGB}{0,100,0}

\title{Enrich-on-Graph: Query-Graph Alignment for Complex Reasoning with LLM Enriching}

\author{
    Songze Li\textsuperscript{1,3},
    Zhiqiang Liu\textsuperscript{1,3},
    Zhengke Gui\textsuperscript{2}, 
    Huajun Chen\textsuperscript{1,3},
    \textbf{Wen Zhang\textsuperscript{1,3}\thanks{~~Corresponding authors.}}\\
    \textsuperscript{1}Zhejiang University,
    \textsuperscript{2}Ant Group,
    \textsuperscript{3}ZJU-Ant Group Joint Lab of Knowledge Graph \\
    \texttt{
    \{li.songze,zhang.wen,huajunsir\}@zju.edu.cn
    }
}

\begin{document}
\maketitle
\begin{abstract}
Large Language Models (LLMs) exhibit strong reasoning capabilities in complex tasks. However, they still struggle with hallucinations and factual errors in knowledge-intensive scenarios like knowledge graph question answering (KGQA). We attribute this to the semantic gap between structured knowledge graphs (KGs) and unstructured queries, caused by inherent differences in their focuses and structures. Existing methods usually employ resource-intensive, non-scalable workflows reasoning on vanilla KGs, but overlook this gap. To address this challenge, we propose a flexible framework, Enrich-on-Graph (EoG), which leverages LLMs' prior knowledge to enrich KGs, bridge the semantic gap between graphs and queries. EoG enables efficient evidence extraction from KGs for precise and robust reasoning, while ensuring low computational costs, scalability, and adaptability across different methods. Furthermore, we propose three graph quality evaluation metrics to analyze query-graph alignment in KGQA task, supported by theoretical validation of our optimization objectives. Extensive experiments on two KGQA benchmark datasets indicate that EoG can effectively generate high-quality KGs and achieve the state-of-the-art performance. Our code and data are available at \url{https://github.com/zjukg/Enrich-on-Graph}.
\end{abstract}

\section{Introduction}

\begin{figure}[h!]
    \centering
    \includegraphics[width=\linewidth, keepaspectratio]{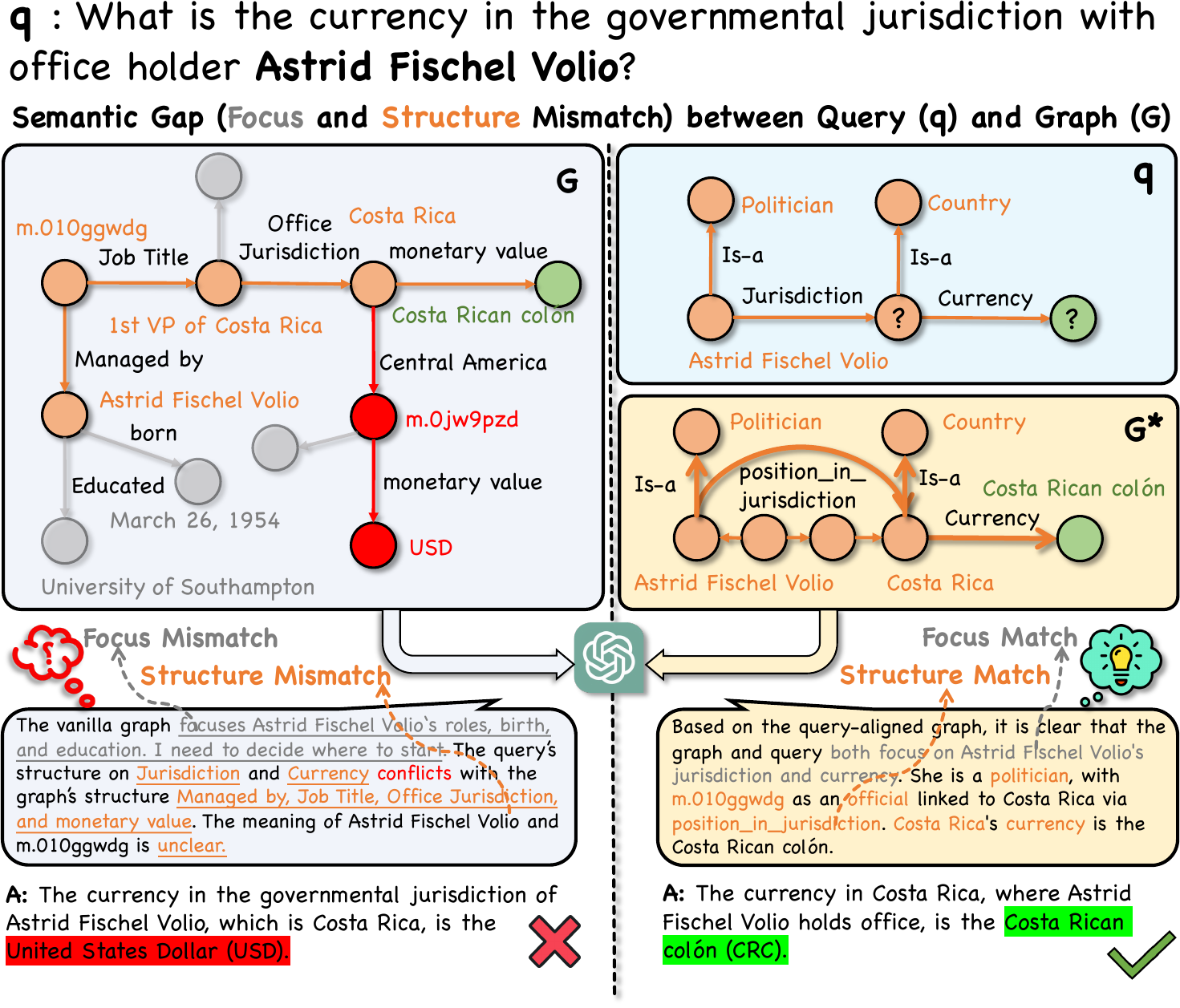} 
    \caption{Semantic gap between query and graph: Gray indicates noise, red represents errors, orange denotes reasoning-related information, and green is answer. We use $ G $, $ Q $, $ G^* $ to represent vanilla graph, query, query-aligned graph, respectively, and use their logic forms for illustration.
\textbf{Left:} LLMs misextracts key information due to the semantic gap between \(Q\) and \(G\).
\textbf{Right:} EoG generates \(G^*\) for efficient LLM reasoning.}
    \label{fig:intro1}
\end{figure}


Large language models (LLMs)~\cite{openai2024gpt4technicalreport,brown2020languagemodelsfewshotlearners,chowdhery2022palmscalinglanguagemodeling,azaria2024chat} excel in complex natural language processing (NLP) tasks~\cite{wei2023chainofthoughtpromptingelicitsreasoning,khot2023decomposedpromptingmodularapproach,li2023evaluation,LiuZZFFZHZD26} due to extensive pre-training on large corpora~\cite{rawte2023surveyhallucinationlargefoundation} embedding prior knowledge in their parameters~\cite{khot2023decomposedpromptingmodularapproach}.
Based on the prior knowledge, LLMs can achieve semantic understanding, perform reasoning, and generate reasonable responses in diverse question-answering tasks~\cite{li2023unigenunifiedgenerativeframework,zhao2024generating,ye-etal-2026-tableqa,guo-etal-2026-rethinking-table,ye2026rethinkingstepwisemodelrouting,guo-etal-2025-sqlforge}. However, LLMs still face challenges like hallucinations and factual errors~\cite{Ji_2023,chen2024large}, particularly in knowledge-intensive scenarios like knowledge graph question answering. 

Knowledge graph question answering (KGQA) is the task of answering natural language queries based on structured factual information stored in knowledge graphs (KGs)~\cite{DBLP:conf/semweb/AuerBKLCI07,freebase}. Existing KGQA methods can be broadly categorized into information retrieval-based and semantic parsing-based.
Information retrieval~\cite{sun-etal-2019-pullnet,zhang-etal-2022-subgraph} methods extract subgraphs relevant to a query and reason over them, but retrieval process inevitably introduces noise, thereby reducing accuracy of answer. Semantic parsing methods~\cite{sun2020sparqaskeletonbasedsemanticparsing,jiang2023unikgqa} generate logical forms (e.g., SPARQL) for querying KGs. 
However, both approaches remain limited by the reasoning capabilities of their underlying models, especially for complex queries.

Given LLMs' strong understanding and reasoning capabilities, advanced methods usually employ LLMs for KGQA. 
\textit{We believe that the core of such methods lies in how to bridge the \textbf{Semantic Gap Between Queries and Knowledge Graphs}}, which stems from the focus and structure mismatch. User query is a precise, goal-driven request with clear semantic focus. In contrast, KGs encompass diverse focuses across many topics, often containing substantial noisy information. This focus mismatches between the query and the KG makes it challenging to accurately retrieve the subgraphs needed for reasoning. Even when relevant subgraphs are retrieved, the rigid structure of KGs often clashes with the linguistic diversity of user queries, complicating reasoning.
As shown in Fig.~\ref{fig:intro1} Left, the vanilla graph contains significant noise (gray entities), which is mismatched with the query focus. The query, \textit{``What is the currency in the governmental jurisdiction with office holder Astrid Fischel Volio?"}, requires a two-hop reasoning path (\textit{Jurisdiction → Currency}), but the vanilla graph uses a 4-hop path (\textit{Managed → Job → Office Jurisdiction → Monetary Value}) and involves ambiguous entity \textit{Astrid Fischel Volio}, creating a structure mismatch.
Consequently, reasoning over vanilla KGs is hindered by focus and structure mismatches. 

\textbf{Previous methods struggle to align the semantics between queries and graphs from the perspective of designing complex reasoning pipelines}, which can be effective at times but come with high computational or training costs that hinder their efficiency.
For example, the DoG~\cite{ma2025debate} framework iteratively simplifies queries and focuses on subgraphs through steps like Invoking, Filtering, Answer Trying, and Simplifying, enabling step-by-step reasoning but leading to a rigid and bulky workflow. RoG~\cite{luo2024reasoninggraphsfaithfulinterpretable} requires fine-tuning LLMs for relationship path planning, which extracts query-relevant subgraph relations for faithful reasoning, but making adaptation to different KGs costly due to retraining. ToG~\cite{sun2023thinkongraph} leverages LLMs to explore and reason over entities and relations in KG based on the query in an iterative manner. 
These methods attempt to align the semantics between queries and graphs through reasoning, but they still suffer from the semantic gap when reasoning over vanilla KGs.

To address the challenge, we propose Enrich-on-Graph (EoG), a flexible three-stage framework that leverages LLMs' prior knowledge to enrich graph, aligning the semantics between the vanilla KGs and queries. 
EoG proceeds in three stages: 
(1) \textbf{Parsing:} parsing the query and graph to enable effective alignment;
(2) \textbf{Pruning:} proposing focus-aware multi-channel pruning to mitigate focus mismatches;
(3) \textbf{Enriching:} leveraging LLMs to enrich graph to resolve structure mismatches.
The aligned graph then collaborates with LLMs as an external knowledge source for efficient reasoning, shown in Fig.~\ref{fig:intro1} Right. 
We also introduce three graph quality metrics—\textbf{Relevance, Semantic Richness}, and \textbf{Redundancy}—to evaluate graphs' structural properties, feature attributes, and LLM-driven aspects, supported by theoretical analyses further linking these metrics to optimization objectives.
The major contributions of this work are as follows:

\begin{itemize}
\item We highlight the key of KGQA task lies in how to bridge the semantic gap between queries and KGs. We further propose three graph quality metrics—Relevance, Semantic Richness, and Redundancy. Theoretical analyses validate alignment mechanism's effectiveness and the practical utility of these metrics.

\item We propose a flexible Enrich-on-Graph (EoG) framework that leverages LLMs as prior to align the semantics between queries and graphs, enabling high-quality KG generation and precise reasoning.

\item Extensive experiments on two KGQA benchmark datasets demonstrate that EoG achieves the state-of-the-art performance through generating high-quality KGs, while ensuring flexibility, low computational cost, scalability, and broad compatibility with other methods.
\end{itemize}

\section{Task Formulation and Analysis}
\label{sec:methods}
\subsection{Task Formulation}

Given a KG $G = \left\{ \left( {e_{s},r,e_{o}} \right) \middle| e_{s},e_{o}\in E,r\in R \right\}$, where $E$ and $R$ are the set of entities and relations, it stores a large amount of factual knowledge in the form of triples. For a complex query $q$, the goal of the KGQA method $M_{\theta}$ with parameters $\theta$ is to get the correct answer $a^*$ that $a^* = M_{\theta}(q, G)$. 

\subsection{Semantic Gap between Query and Graph}
\label{Semantic Gap between Query and Graph}

\begin{figure}[h!]
    \centering
    \includegraphics[width=\linewidth, keepaspectratio]{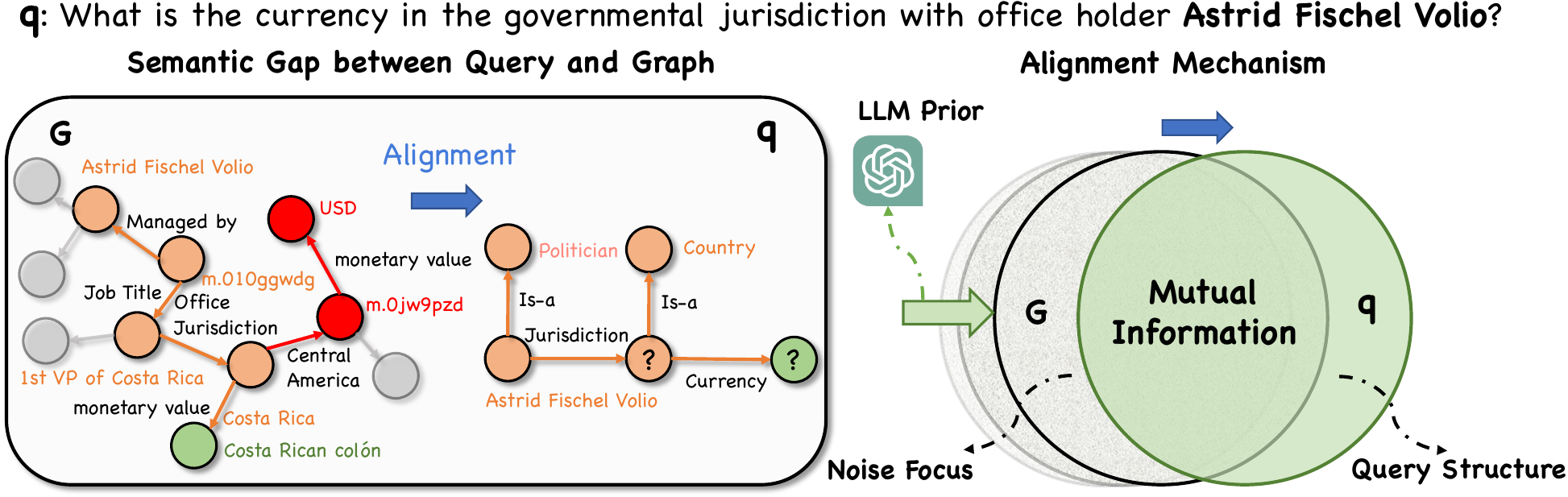}
    \caption{\textbf{Left:} focus mismatch and structure mismatch between query and vanilla graph cause semantic gaps.
\textbf{Right:} Demonstration of EoG's alignment mechanism.}
    \label{fig:method1}
\end{figure}

To better illustrate that the semantic gap between the KG and the query arises from \textbf{focus mismatch} and \textbf{structure mismatch}, we refer to previous works~\cite{wang2023knowledgedrivencotexploringfaithful,DBLP:conf/aaai/ZhangZMWS0SZ24} to visualize their logical forms, as shown in Fig.~\ref{fig:method1} (left).
\paragraph{Focus Mismatch:} For the query \textit{``What is the currency in the governmental jurisdiction with office holder Astrid Fischel Volio?"}, it explicitly focuses on the jurisdiction managed by \textit{Astrid Fischel Volio} and its currency, represented as: \textit{Astrid Fischel Volio → Jurisdiction → ? → Currency → ?}. However, the vanilla KG includes irrelevant focuses, such as \textit{Astrid Fischel Volio}’s birthday and education (gray nodes), which we call focus mismatch since they do not help the reasoning process.
\paragraph{Structure Mismatch:} As shown in Fig.~\ref{fig:method1} (left), the vanilla KG contains 4-hop structures (orange nodes): \textit{Astrid Fischel Volio → Managed → m.010ggwdg → Job → 1st VP of Costa Rica → Office Jurisdiction → Costa Rica → Monetary Value → Costa Rican colón}. In contrast, the query has corresponding 2-hop structures: \textit{Jurisdiction → Currency}. The \textbf{structure inconsistency in reasoning hops} (e.g., the query’s single-hop \textit{Jurisdiction} versus KG’s 3-hop \textit{Managed → Job → Office Jurisdiction}) makes multi-hop reasoning harder for LLMs. Additionally, ambiguous entities like \textit{Astrid Fischel Volio} (\textbf{lacking a hierarchy structure} such as \textit{Astrid Fischel Volio → is-a → Politician}) further confuse LLMs. These issues are referred to as structure mismatch.
In summary, \textbf{the semantic gap between the vanilla KG and the query} stems from \textbf{focus mismatch} and \textbf{structure mismatch}.

\section{Method}
\subsection{Solution of EoG}
As shown in Fig.~\ref{fig:method1} (left), ideally, for complex KGQA tasks, we want to use a graph $G^*$ that is consistent with the logic form of $q$ to help LLMs reason efficiently. However, in reality, we can only use the vanilla KG $G$, letting LLMs leverage external knowledge from $G$ and its pre-trained knowledge to find information relevant to $q$ for answering. Thus, our goal is to find an optimized graph $G^*$ by maximizing the expected posterior probability:

\[{G^{*} = \underset{G}{argmax}}{~\mathbb{E}_{P(q,G)}\left\lbrack P\left( M_{\theta},q \middle| G \right) \right\rbrack}\]
By identifying the optimized graph $  G^*  $, the LLM $  M_{\theta}  $ can directly extract key knowledge from $  G^*  $ to better perform reasoning on $q$ and arrive at $a^*$.
\begin{theorem}
Maximizing the expected posterior probability is equivalent to maximizing the mutual information (MI) between $q$ and $G$.
\end{theorem}
\label{theorem1}

{\small
\[P\left( M_{\theta},q \middle| G \right) \propto {\log\left( \frac{P\left( M_{\theta},q \middle| G \right)}{P(q)} \right)}{{= \log}\left( \frac{P\left( M_{\theta},G \middle| q \right)}{P(G)} \right)}\]
}
{\small
\[\propto {\log\left( \frac{\int_{M_{\theta}}^{}{P\left( M_{\theta},G \middle| q \right)dM_{\theta}}}{P(G)} \right)} = {\log\left( \frac{P\left( {q,G} \right)}{P(q)P(G)} \right)}\]
}
Therefore, we can obtain (Details in Appendix~\ref{Details of our Theoretical Proof}):
\[\mathbb{E}_{P(q,G)}\left\lbrack {P\left( M_{\theta},q \middle| G \right)} \right\rbrack \propto MI(q,G)\]

As shown in Fig.~\ref{fig:method1} (right), through maximizing $MI(q, G)$, $G$ eliminates focus mismatch and structure mismatch, resulting in the desired $G^*$.

\subsection{Our Method: Enrich-on-Graph (EoG)}
\begin{figure*}[ht]
    \centering
    \includegraphics[width=1.0\textwidth, keepaspectratio]{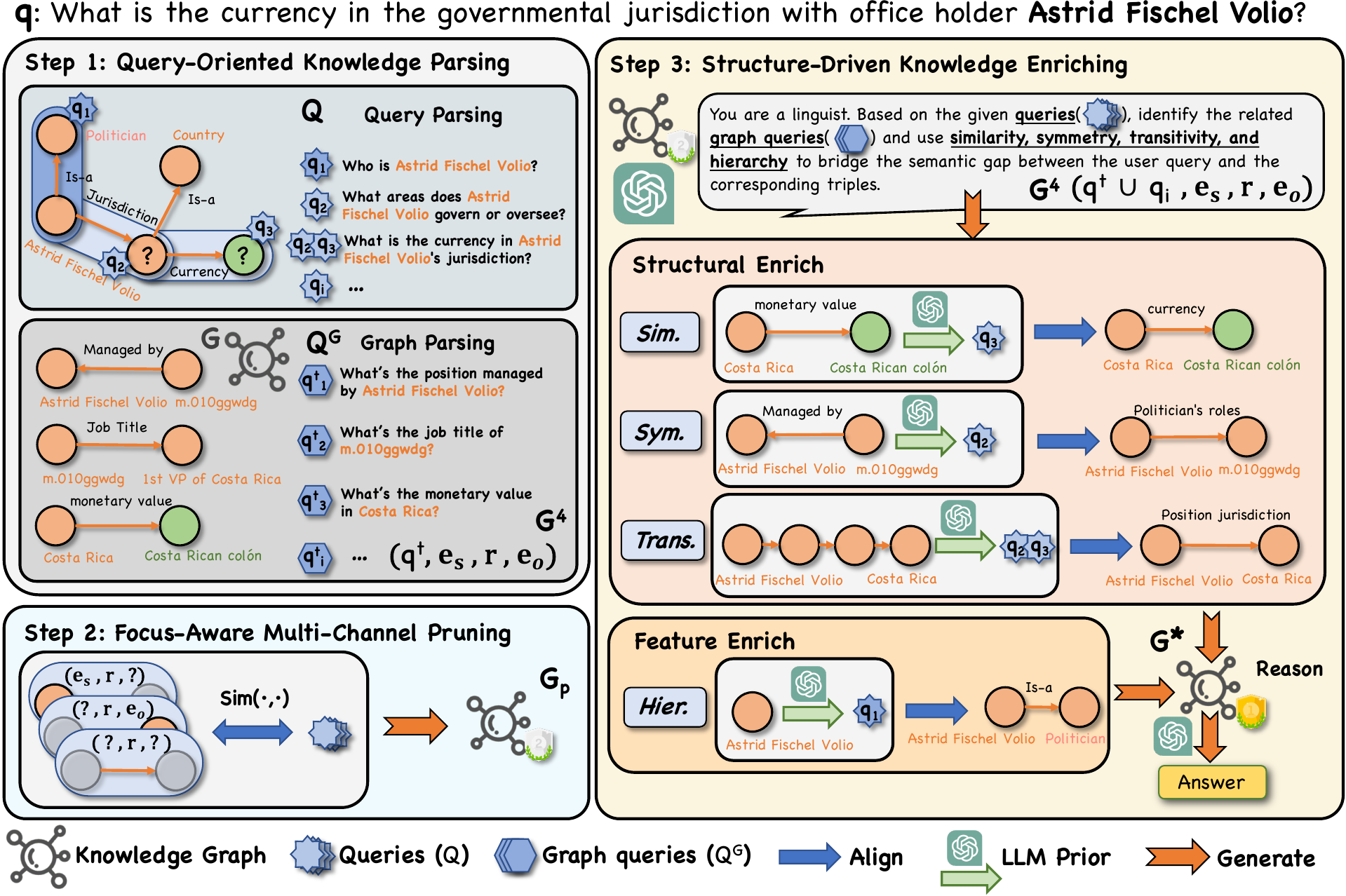}
    \caption{Overview of our Enrich-on-Graph framework.}
    \vspace{-4mm}
    \label{fig:method2}
\end{figure*}

As shown in Fig.~\ref{fig:method2}, given a user query $q$ and a vanilla graph $G$, our method consists of three stages:
\textbf{(1) Parsing:} we parse $q$ into new forms $Q$, while converting $G$ into a query form $Q^G$ (which we refer to as the graph query) and injecting it into triples to construct quadruples, preparing for the better semantic alignment.
\textbf{(2) Pruning:} Using the $Q$ from the first stage, we prune noisy graph focuses via multiple channels, eliminating focus mismatches.
\textbf{(3) Enriching:} Leveraging LLMs with parametric knowledge as prior, $Q^G$ related to the $Q$ are identified and corresponding structures in quadruples are semantically aligned using KG's structural and feature properties, thus enhancing reasoning.

\subsubsection{Query-Oriented Knowledge Parsing}
\label{step 1}
The complex query $q$ is in natural language, while $G$ is in triple form, making their formats different and hindering subsequent semantic alignment. To address this, we use LLMs to convert $q$ and $G$ into the same query format, $Q$ and $Q^G$.

\paragraph{Query Parsing:} LLM breaks down $q$ into sub-queries, which are categorized into two types: compound queries (requiring multi-hop reasoning) and unit queries (single-hop reasoning), collectively forming \(Q=\{q,q_1,q_2,...\}\) (Appendix~\ref{Query Structuring Prompt} for prompts). For example, the $q$ \textit{``What is the currency in the governmental jurisdiction with office holder Astrid Fischel Volio?"} is decomposed into a compound query $q_1$ \textit{``What is the currency in Astrid Fischel Volio's jurisdiction?"} and unit queries like $q_2$ \textit{``What areas does Astrid Fischel Volio oversee?"} and $q_3$ \textit{``Who is Astrid Fischel Volio?"}. 

\paragraph{Graph Parsing:} LLM transforms triple $t$ into graph query $q^t$ (all triples' graph queries $q^t$ collectively form the set $Q^G$) to construct quadruples $G^4 = (q^t, e_s, r, e_o)$, preserving the original knowledge graph while introducing graph queries for the subsequent alignment of graphs and query. For instance, the triple \textit{(Costa Rica, monetary value, Costa Rican colón)} is converted into the graph query: \textit{``What’s the monetary value in Costa Rica?"} to construct (\textit{``What’s the monetary value in Costa Rica?", Costa Rica, monetary value, Costa Rican colón}).
$q^t$ and $q_i$ ($q_i \in Q$) are in the same query format to facilitate subsequent alignment.

\subsubsection{Focus-Aware Multi-Channel Pruning}
\label{Structure-Aware Multi-Channel Pruning}

To address the focus mismatch between $q$ and $G$, we adopt pruning techiques to remove noisy focuses in $G$, inspired by previous approaches (HOMELS~\cite{panda2024holmeshyperrelationalknowledgegraphs}, QA-GNN~\cite{yasunaga2022qagnnreasoninglanguagemodels}, and DynaGraph~\cite{10.1145/3534540.3534691}). However, these naive pruning methods only focus $q$ and $G$ at a global level, simply pruning based on query-triple similarity, while ignoring local focuses, which often introduces noise.

To solve this, we propose \textbf{Focus-Aware Multi-Channel Pruning} that considers both global and local focuses. The $Q$ from~\ref{step 1} contains global and local focuses: compound queries (requiring multi-hop reasoning) contain more global information, representing the query's global focuses, while unit queries only require single-hop reasoning, representing the query's local focuses.

For each triple $t = (e_s, r, e_o)$ in $G$, we mask the head entity ($MASK_1(t) = (?, r, e_o)$), tail entity ($MASK_2(t) = (e_o, r, ?)$), and both ($MASK_3(t) = (? , r, ?)$) to create three recall channels, capturing various local focuses. 

{\small
\[G_{p} = Top~K\left( {\sum_{j}{\sum_{i}{sim\left( q_{i},{MASK}_{j}(t) \right)}}} \right),\]
}

where $q_{i} \in Q, t\in {G}$. To fully focus the semantics between $q$ and $G$ from both global and local perspectives during pruning, we compute semantic similarity between $Q$ and each triple of $G$ in three recall channels. For example, in the $MASK_1$ channel, we calculate the dot-product similarity between all queries in $Q$ and each masked triple $(?, r, e_o)$, summing the scores. The final score of a triple combines scores across all recall channels. Elaboration of Focus-Aware Multi-Channel Pruning is in Appendix \ref{elaboration-Focus-Aware Multi-Channel Pruning}. We keep the top $K$ triples $t_p$ with the highest final scores as the pruned graph $G_p$ (Appendix~\ref{Optimal k-value Exploration} details ablation studies on $K$).


\subsubsection{Structure-Driven Knowledge Enriching}
\label{Structure-Driven Knowledge Enrichment}
To address the \textbf{structure mismatch} caused by \textbf{structure inconsistency in reasoning hops} and \textbf{lacking hierarchy structure} (see~\ref{Semantic Gap between Query and Graph}), we propose \textbf{Structure-Driven Knowledge Enriching} to optimize the structures of $G_p$.
First, to efficiently align the structures of $q$ and $G_p$, we use the LLM to filter the relevant $q^{t_p}$ from $G^4$ for each $q_i$ in $Q$ (obtained in~\ref{step 1}), and then integrate $q_i$ into the quadruples $G^4 = (q_i \cup q^{t_p}, e_s, r, e_o)$. Since both are parsed into the same query format, LLM can easily handle this integration.

Inspired by previous work~\cite{han2025retrievalaugmentedgenerationgraphsgraphrag, article2}, KGs have two key properties: structural attributes (similarity, symmetry, and transitivity of entity relationships) and feature attributes (hierarchical ontologies). Based on these, we improve the reasoning paths in $G_p$ (both single-hop and multi-hop structures) using structural properties and enrich entity hierarchy structures in $G_p$ with feature properties to mitigate ambiguity.

Specifically, we use $Q$ as indicators to enrich graph structures $(e_s, r, e_o)$ in $G^4$.
For single-hop structures, \textbf{similarity} and \textbf{symmetry} complement semantics. For example, for quadruples \textit{(``What areas does Astrid Fischel Volio govern or oversee?", Astrid Fischel Volio, Managed by, m.010ggwdg)}, LLM generates the reverse relation \textit{Politician's roles} for \textit{Astrid Fischel Volio}, aligning better with $q$.
For multi-hop structures, \textbf{transitivity} reduces reasoning complexity. The 3-hop structure from \textit{Astrid Fischel Volio} to \textit{Costa Rica} is simplified to a direct relation through \textit{Position jurisdiction}, aligning more closely with $q$ and streamlining reasoning.
Lastly, \textbf{hierarchy} enriches hierarchy structures of entities $ e_s $ and $ e_o $ by generating query-related ontology \cite{liang2026oc}, reducing ambiguity and aiding precise reasoning. For example, for the entity \textit{Astrid Fischel Volio}, its ontology triple \textit{(Astrid Fischel Volio, is-a, Politician)} clarifies identity. The enriched graph is then used to assist LLM in question answering efficiently. 
(How EoG's aligned graphs help LLMs reason efficiently is detailed in Appendix~\ref{Case Study}. Enriching and Question Answering prompts are in the Appendix~\ref{Structural Enrich Prompt},~\ref{Feature Enrich Prompt},~\ref{Question Answering Prompt}.)

\subsection{Graph Quality Evaluation Metrics}
To verify that our proposed method facilitates progress toward the optimization objective, we design graph evaluation metrics and theoretically prove their correlation with the optimization goal. 

\paragraph{Relevance:} As discussed in~\ref{Structure-Driven Knowledge Enrichment}, KGs have structural and feature-based properties. For structural property, we aim to eliminate noisy focuses and address structure inconsistency in reasoning hops between the $q$ and $G$. To measure relevance, we define a metric where $v_q$ and $v_t$ are the embeddings of query $q$ and triple $t$, respectively.

{\small
\[S^{r}(q,G) = {\sum\limits_{t \in G}{sim\left( v_{q},v_{t} \right)}}\]
}

\paragraph{Semantic Richness:} For the feature property of KG, we need to enirch the hierachy structures of entities in $G$ to address the ambiguity issues, providing more semantically enriched ontological information. Therefore, we have designed a semantic richness metric:

{\small
\[S^{e}(G) = ~{\sum\limits_{t \in G}{KGC\left( t_{+} \right)}},\]
}
where $ t_{+} $ indicates positive triples, $KGC$ is the semantic scoring model, such as KG-BERT~\cite{yao2019kgbertbertknowledgegraph}, which evaluates the completeness score of the triples in the semantic space.

\paragraph{Redundancy:} For the KGQA task, redundant and repetitive triples are meaningless, as they do not provide additional effective information for LLM reasoning but instead increase computational overhead~\cite{liao-etal-2025-skintern,liu-etal-2024-lost,yuan2024hybridragcomprehensiveenhancement}. Therefore, we design a redundancy metric.

{\small
\[S^{d}(G) = {\sum\limits_{G^{sub} \in G}{{\sum\limits_{r_{j_{1}} \in G^{sub}(r)}{\sum\limits_{r_{j_{2}} \in G^{sub}(r)}{sim\left( {v_{r_{j_{1}}},v_{r_{j_{2}}}} \right)}}},}}\]
}
{\small
\textit{
\[{j_{1} \neq j_{2},G}^{sub} = \left\{ \left( e_{s},r,e_{o} \right) \middle| e_{s} = e_{s}^{fixed},e_{o} = e_{o}^{fixed} \right\}\]
}
}

Here, $G(r)$ extracts the relationship in the triples, $e_{s}^{fixed}$ and $e_{o}^{fixed}$ denote subgraphs consisting of triples with the same head entity and tail entity, respectively.
\begin{theorem}
The relationship between the graph quality evaluation metrics Relevance and Semantic Richness are positively correlated with the optimization objective $  MI(Q, G)  $.
\end{theorem}

{\tiny
\[P\left( {q,G} \right) = \frac{\sum\limits_{j = 1}^{l}{n\left( q,t_{j} \right)}}{l} \propto \frac{\sum\limits_{j = 1}^{l}{sim\left( v_{q},v_{t_{j}} \right)}}{l} \]
}
{\tiny
\[P\left( {q,G} \right) = \frac{\sum\limits_{j = 1}^{l}{n\left( {q,t_{j}} \right)}}{l} \propto \frac{\sum\limits_{t_{+} \in G}{n\left( q,t_{+} \right)}}{l} \propto \frac{\sum\limits_{t_{+} \in G}{KGC\left( t_{+} \right)}}{l}\]
}

Therefore, we can obtain:

{\small
\[MI(q,G)~ \propto S^{r},S^{e}\]
}
We prove that maximizing $MI(q, G)$ using LLM prior knowledge aligns the semantics of $q$ and $G$, improving graph quality metrics. Experiments further validate that EoG generates higher-quality graphs, contributing to the achievement of the optimization goal.

\begin{table}
\centering
\resizebox{0.85\linewidth}{!}{
\begin{tabular}{l|cccc}
\toprule
\multicolumn{1}{l|}{\multirow{2}{*}{Model}} & \multicolumn{2}{c}{CWQ}  & \multicolumn{2}{c}{WebQSP}  \\ 
 & \textit{Hit@1} & \textit{F1} & \textit{Hit@1} & \textit{F1} \\ \midrule
\multicolumn{5}{c}{\textit{Information Retrieval}} \\ \midrule
KV-Mem & 21.1 & 15.7 & 46.7 & 34.5 \\
GraftNet & 36.8 & 32.7 & 66.4 & 60.4 \\
PullNet & 47.2 & - & 68.1 & - \\
EmbedKGQA & 44.7 & - & 66.6 & - \\
NSM+h & 48.8 & 44.0 & 74.3 & 67.4 \\
TransferNet & 48.6 & - & 71.4 & - \\
Subgraph Retrieval & 50.2 & 47.1 & 69.5 & 64.1 \\ \midrule
\multicolumn{5}{c}{\textit{Semantic Parsing}} \\ \midrule
SPARQL & 31.6 & - & - & - \\
UHop & - & 29.8 & - & 68.5 \\
Topic Units & 39.3 & 36.5 & 68.2 & 67.9 \\
QGG & 44.1 & 40.4 & 73.0 & 74.0 \\
UniKGQA & 51.2 & 49.4 & 77.2 & 72.2 \\
TIARA & - & - & 75.2 & - \\ \midrule
\multicolumn{5}{c}{\textit{LLMs}} \\ \midrule
Flan-T5-xl & 14.7 & - & 31.0 & - \\
Alpaca-7B & 27.4 & - & 51.8 & - \\
LLaMa2-Chat-7B & 34.6 & - & 64.4 & - \\
IO prompt w/ ChatGPT & 37.6 & - & 63.3 & - \\
CoT prompt w/ ChatGPT & 38.8 & - & 62.2 & - \\
SC prompt w/ ChatGPT & 45.4 & - & 61.1 & - \\
InteractiveKBQA & 59.2 & - & 72.5 & - \\ \midrule
\multicolumn{5}{c}{\textit{LLMs+KGs}} \\ \midrule
StructGPT & 54.3 & - & 72.6 & - \\
KD-CoT & 50.5 & - & 73.7 & 50.2 \\
DeCAF & 70.4 & - & 82.1 & - \\
KG-CoT & 62.3 & - & 84.9 & - \\
RoG & 62.6 & 56.2 & 85.7 & 70.8 \\
ToG & 69.5 & - & 82.6 & - \\
DoG & 58.2 & - & \textbf{91.0} & - \\
\textbf{EoG (Ours)} & \textbf{70.8} & \textbf{65.1} & 85.0 & \textbf{74.1} \\ \bottomrule
\end{tabular}
}
\caption{Performance comparison of EoG and various baselines on CWQ and WebQSP datasets, with the best results in bold.}
\label{main result}
\end{table}

\section{Experiment}
\subsection{Experimental Setup}
\paragraph{Datasets.}We evaluate reasoning performance on benchmark KGQA datasets used by: WebQSP~\cite{DBLP:conf/acl/YihRMCS16} includes 4,737 questions involving simple and two-hop reasoning, while CWQ~\cite{Talmor2018TheWA} features 34,689 questions requiring complex 2-4 hop reasoning. Both datasets use Freebase~\cite{Bollacker2008FreebaseAC} as the KG, with 88 million entities, 20,000 relations, and 126 million triples. More details are in the Appendix~\ref{datastes statics}.
\paragraph{Baselines.} We compared KGQA baselines from Section~\ref{related work}: information retrieval, semantic parsing, LLM reasoning with and without KGs.
\paragraph{Evaluation Metrics.}
Following previous work, we use Hits@1 for top-answer accuracy and F1 to assess coverage for multi-answer questions, balancing Precision and Recall.
\paragraph{Implementations.}During the pruning stage, we use sentence-transformers~\cite{reimers2019sentencebertsentenceembeddingsusing} as the retriever, setting the top $k = 300$. GPT-4o mini serves as our base model, with temperature set to 0.2, generation count $n$ and sampling parameter top $p$ set to 1, and max tokens set to the model’s maximum output length for reproducibility.
\subsection{Main Result}

Tab.~\ref{main result} compares the performance of our proposed EoG method with existing SOTA approaches. The LLM for KG method achieves superior performance compared to other mainstream models. Our approach outperforms other LLM-based KGQA methods across nearly all metrics on the CWQ and WebQSP datasets. On WebQSP, we improve F1 by 4.7\% over the advanced RoG. On the challenging CWQ dataset, our method demonstrates overall superiority, with Hits@1 and F1 improved by 13.1\% and 15.8\%, respectively, compared to RoG, and Hits@1 improved by 4.7\% and 0.6\% compared to ToG and DeCAF. These results highlight the SOTA performance of our method on KGQA tasks.

\subsection{Abalation Study}
\begin{table}[]
\centering
\resizebox{\linewidth}{!}{
\begin{tabular}{l|cccc|cccc}
\toprule
\multirow{2}{*}{Model} & \multicolumn{4}{c}{CWQ} & \multicolumn{4}{|c}{WebQSP} \\ 
\multicolumn{1}{c|}{} & \textit{Acc.} & \textit{Hit@1} & \textit{F1} & \textit{Prec.} & \textit{Acc.} & \textit{Hit@1} & \textit{F1} & \textit{Prec.} \\ \midrule
w/o $\mathds{E}_s$ & 62.5 & 68.3 & 61.5 & 66.6 & 71.8 & 84.1 & 72.5 & 86.3 \\
w/o $\mathds{E}_f$ & 64.8 & 70.2 & 64.2 & 69.2 & 72.5 & 84.5 & 73.7 & 87.6 \\ 
w/o $\mathds{E}_o$ & 52.4 & 58.8 & 50.9 & 56.1 & 66.2 & 81.4 & 67.2 & 83.0 \\
w/o $\mathds{P}$ \& $\mathds{E}_o$ & 49.8 & 55.6 & 47.9 & 52.1 & 64.1 & 79.8 & 64.1 & 77.7 \\ \midrule
\textbf{EoG} & \textbf{65.5} & \textbf{70.8} & \textbf{65.1} & \textbf{70.4} & \textbf{73.0} & \textbf{85.0} & \textbf{74.1} & \textbf{95.6} \\  \bottomrule
\end{tabular}
}
\caption{Ablation study of EoG's core modules, comparing the results after removing the Enrich and Prune modules separately.}
\label{abalation study}
\end{table}

We conducted ablation studies to assess the effectiveness of the Prune and Enrich modules, summarized in Tab.~\ref{abalation study}. The Enrich module includes Structural and Feature submodules, examined under four settings: (1) w/o Prune ($\mathds{P}$), (2) w/o Enrich ($\mathds{E}_o$), (3) w/o Feature Enrich ($\mathds{E}_f$), and (4) w/o Structural Enrich ($\mathds{E}_s$).
\textbf{Alleviating focus mismatches:} The Prune module improves all performance metrics by reducing significant noise in vanilla KG.
\textbf{Mitigating structure mismatches:} 
Using Structural Enrich alone increases Hits@1 and F1 by 3.8\% and 9.7\% on WebQSP, and 19.4\% and 26.1\% on CWQ. Feature Enrich alone boosts Hits@1 and F1 by 3.3\% and 7.9\% on WebQSP, and 16.2\% and 21\% on CWQ. Combining both submodules achieves even better results than using either individually.
These findings demonstrate the effectiveness of the Prune and Enrich modules in query-graph alignment.
\subsection{Graph Evaluation Metrics}

\begin{figure}[h!]
    \centering
    \includegraphics[width=\linewidth, keepaspectratio]{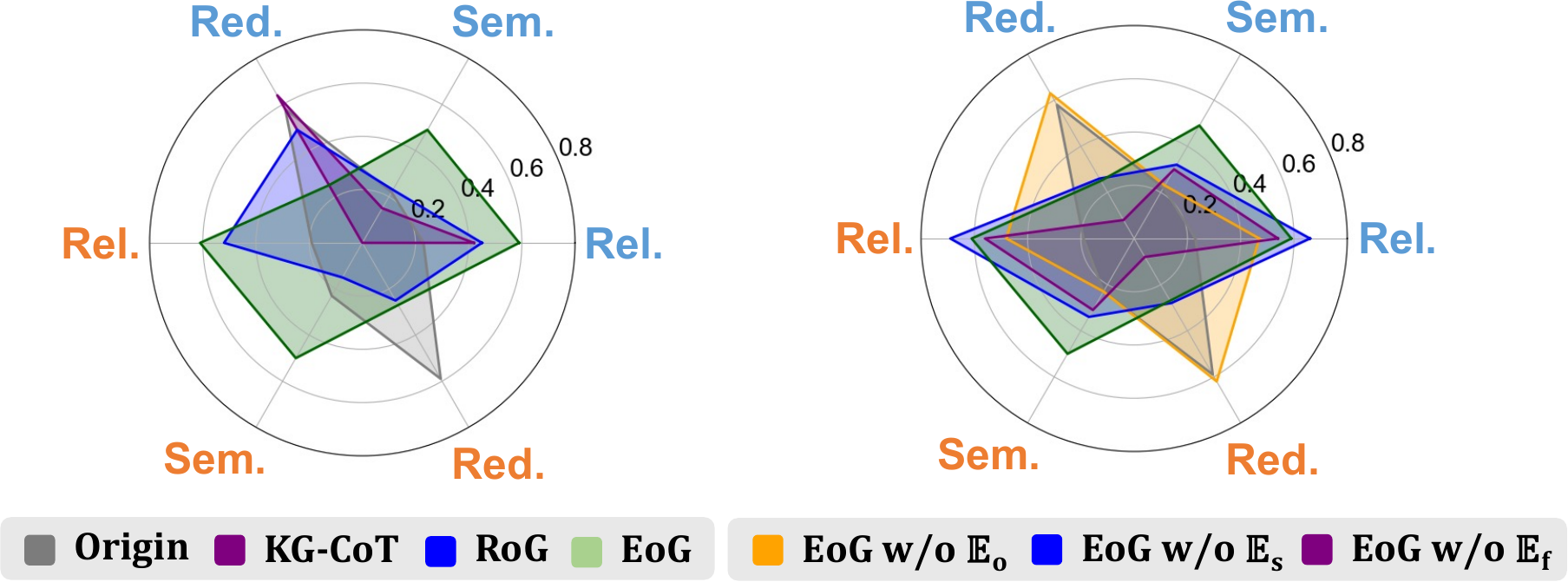}
    \caption{\textbf{Left:} Comparison of graph quality metrics between EoG and other methods.
\textbf{Right:} Validation of graph quality improvement by Prune and Enrich modules. \textbf{Rel.}, \textbf{Sem.}, and \textbf{Red.} indicate Relevance, Semantic richness, and Redundancy, respectively, and these metrics in the blue and orange respectively represent the results on the CWQ and WebQSP. }
    \label{graph metrics cmp}
\end{figure}


To showcase the advantages of EoG-generated graphs, we evaluated their quality using three metrics: relevance, semantic richness, and redundancy. On the WebQSP and CWQ datasets, EoG was compared with the original dataset, RoG, and KG-CoT. KG-CoT results are only reported for CWQ due to incompatibility with WebQSP (Fig.~\ref{graph metrics cmp} left).
EoG achieves relavance scores above 0.6, outperforming RoG and KG-CoT with less noise. Its semantic richness (\textasciitilde0.5) is 20\% higher, enhancing semantics for better LLM reasoning. Furthermore, EoG has the lowest redundancy (\textasciitilde0.2), ensuring concise graphs with minimal redundant information.
We also analyzed the effects of EoG’s Prune and Enrich modules (Fig.~\ref{graph metrics cmp} right). 
The results show that the Prune improves the relevance score by over 0.25. Structural Enrich and Feature Enrich each increase the semantic richness score by over 0.1, while their combination improves it by over 0.26. Feature Enrich significantly reduces redundancy to below 0.1. This demonstrates the contribution of Prune and Enrich in improving subgraph quality and aligning KGs with the query.

\begin{figure}[ht]
    \centering
    \includegraphics[width=\linewidth, keepaspectratio]{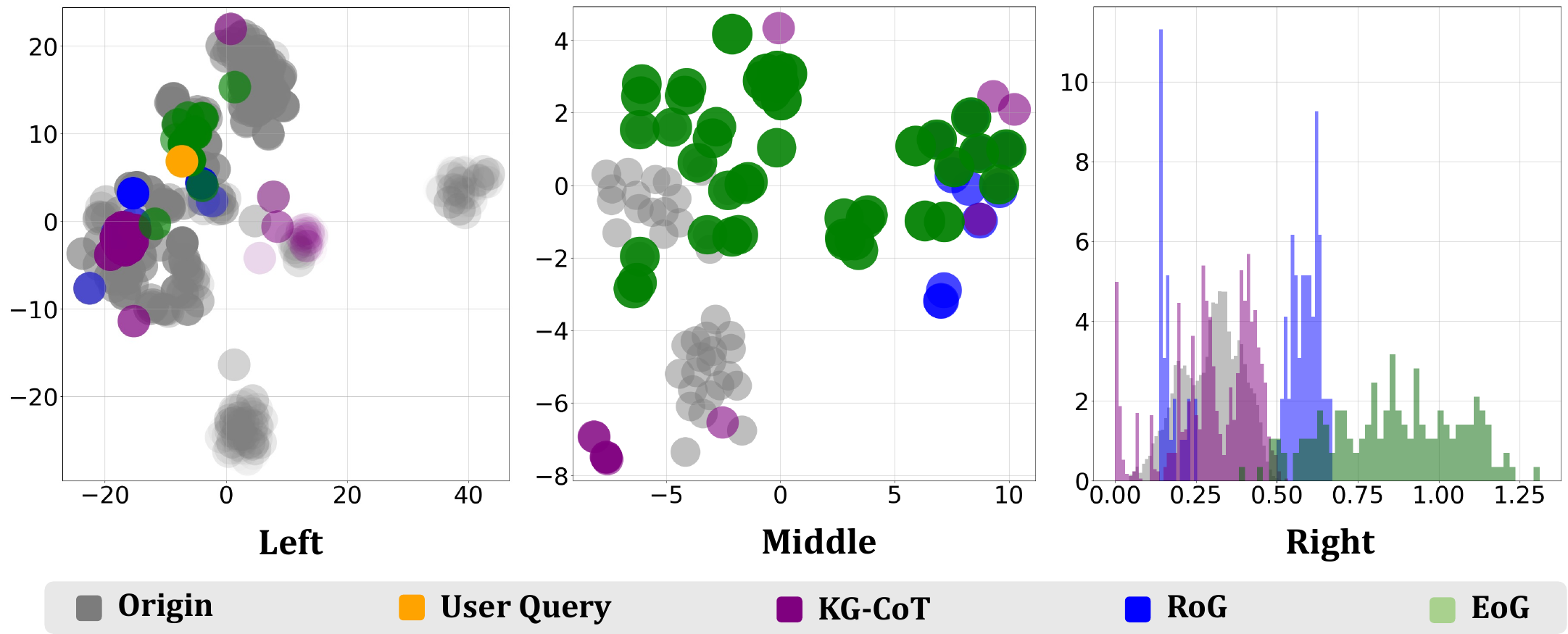}
    \caption{Graph visualization of EoG and advanced methods on the CWQ dataset.
\textbf{Left:} t-SNE projection of \textbf{Relevance}.
\textbf{Middle:} t-SNE visualization of \textbf{Semantic Richness}.
\textbf{Right:} Comparison of \textbf{Redundancy}.}
    \label{fig:cwq graph quality visualization}
\end{figure}

Additionally, we visualized random CWQ cases to illustrate EoG’s graph quality advantages across metrics in Fig.~\ref{fig:cwq graph quality visualization}.
\textbf{Relevance:} Using t-SNE, we plotted triples and question embeddings in 2D, with yellow dots marking question positions. EoG triples are closer and more concentrated around these points compared to the dispersed, noisy clusters of other methods, confirming higher relevance and reduced noise.
\textbf{Semantic Richness:} With t-SNE and KG-BERT semantic scoring, circle sizes represent triple semantic scores (larger circles denote higher scores). EoG covers a significantly larger total area, illustrating richer semantic information.
\textbf{Redundancy:} Triple redundancy was visualized as a density distribution of pairwise Euclidean distances. EoG triples exhibit greater spacing, indicating lower redundancy and more concise graphs. 

\subsection{Computation Cost}
\begin{table}[]
\resizebox{\linewidth}{!}{
\begin{tabular}{l|ccc|ccc}
\toprule
\multicolumn{1}{l|}{\multirow{2}{*}{Model}} & \multicolumn{3}{c|}{CWQ} & \multicolumn{3}{c}{WebQSP} \\ 
\multicolumn{1}{c|}{} & \textit{\# LLM Call} & \textit{Total Token} & \textit{Total Cost} & \textit{\# LLM Call} & \textit{Total Token} & \textit{Total Cost} \\ \midrule
CoT & 1 & 409.7 & 8.00E-05 & 1 & 397.6 & 8.00E-05 \\
ToG & 9.2 & 11468.5 & 2.30E-03 & 8.8 & 10189.4 & 2.10E-03 \\
DoG & 5.7 & 37919.7 & 6.00E-03 & 2.7 & 6114.5 & 1.00E-03 \\ \midrule
EoG & 4 & 6213.6 & 1.10E-03 & 4 & 6802.1 & 1.20E-03 \\
w/o $\mathds{E}_o$ & 2 & 4610.0 & 7.00E-04 & 2 & 4841.6 & 8.00E-04 \\
w/o $\mathds{P}$ \& $\mathds{E}_o$ & 1 & 63814.4 & 9.00E-03 & 1 & 66928.0 & 1.00E-02 \\ \bottomrule
\end{tabular}
}
\caption{Efficiency comparison between EoG and other advanced methods.}
\label{computation cost}
\end{table}
We conducted a computation cost experiment to assess EoG's efficiency by comparing LLM call frequency, token usage, and cost metrics (Tab.~\ref{computation cost}). Results show EoG requires only 4 LLM calls per query, far fewer than ToG and DoG, and reduces token usage by 45.8\% and 83.6\% compared to ToG and DoG, respectively, with the lowest cost. Using the pruning module, EoG further cuts token usage by 92.8\%, highlighting its superior computational efficiency.
\subsection{Plug-and-Play}

\begin{table}[]
\centering
\resizebox{0.8\linewidth}{!}{
\begin{tabular}{l|llll}
\toprule
\multicolumn{1}{l|}{\multirow{2}{*}{Model}} & \multicolumn{2}{c}{CWQ\qquad} & \multicolumn{2}{l}{\quad WebQSP} \\ 
\multicolumn{1}{c|}{} & \;\textit{Hit@1} & \textit{F1} & \textit{Hit@1} & \textit{F1} \\ \midrule
RoG & \;62.2 & 55.4 & 86.4 & 70.8 \\
RoG+$\mathds{E}_o$ & \;\textbf{75.4}\textsubscript{\textcolor{green}{↑13.2}} & \textbf{68.7}\textsubscript{\textcolor{green}{↑13.3}} & \textbf{91.5}\textsubscript{\textcolor{green}{↑5.1}} & \textbf{77.1}\textsubscript{\textcolor{green}{↑6.3}} \\ \midrule
KG-CoT & \;62.3 & 52.7 & \;\,- & \;\,- \\
KG-CoT+$\mathds{E}_o$ & \;\textbf{63.7}\textsubscript{\textcolor{green}{↑1.4}} & \textbf{65.7}\textsubscript{\textcolor{green}{↑13.0}} & \;\,- & \;\,- \\ \midrule
ToG & \;67.6 & \;\,- & 82.6 & \;\,- \\
ToG+$\mathds{E}_o$ & \;\textbf{70.7}\textsubscript{\textcolor{green}{↑3.1}} & \;\,- & \textbf{89.4}\textsubscript{\textcolor{green}{↑6.8}} & \;\,- \\ \midrule
DoG & \;56.0 & \;\,- & 91.0 & \;\,- \\
DoG+$\mathds{E}_o$ & \;\textbf{60.5}\textsubscript{\textcolor{green}{↑4.5}} & \;\,- & \textbf{92.5}\textsubscript{\textcolor{green}{↑1.5}} & \;\,- \\ \bottomrule
\end{tabular}
}
\caption{Impact of the Enrich module as a plugin enhancing other KGQA methods.}
\label{plug-and-play}
\end{table}

To validate EoG's plug-and-play capability, we applied the Enrich module to enhance other KGQA methods. As shown in Tab.~\ref{plug-and-play}, integrating Enrich with RoG, KG-CoT, ToG, and DoG significantly improved Hits@1 and F1, demonstrating its adaptability and robustness across KGQA methods. 
Additionally, experiments on base models with different reasoning capabilities and temperature parameters achieved excellent results, demonstrating EoG's strong reproducibility and plug-and-play flexibility (results in Appendix~\ref{plug-and-play-base-model}).


\section{Related Work}
\label{related work}

\paragraph{Traditional KGQA.}
Traditional KGQA methods are often categorized into information retrieval and semantic parsing approaches. Information retrieval methods (e.g., KV-Mem~\cite{miller-etal-2016-key}, GraftNet~\cite{DBLP:conf/emnlp/SunDZMSC18}, PullNet~\cite{sun-etal-2019-pullnet}, EmbedKGQA~\cite{saxena-etal-2020-improving}, NSM+h~\cite{10.1145/3437963.3441753}, TransferNet~\cite{shi-etal-2021-transfernet}, and Subgraph Retrieval~\cite{zhang-etal-2022-subgraph}) rely on retrieving query-relevant subgraphs from KGs but often retrieve irrelevant noises, leading to focus mismatch and reasoning errors. Semantic parsing (e.g., SPARQL~\cite{sun2020sparqaskeletonbasedsemanticparsing}, UHop~\cite{chen-etal-2019-uhop}, Topic Units~\cite{ijcai2019p701}, QGG~\cite{lan-jiang-2020-query}, UniKGQA~\cite{jiang2023unikgqa}, RnG-KBQA~\cite{ye-etal-2022-rng}, and TIARA~\cite{shu-etal-2022-tiara}) converts user queries into formal representations (e.g., SPARQL) to retrieve answers but struggles with complex queries, causing incomplete or failed answers.
\paragraph{LLMs for KGQA.}\vspace{-5pt}
LLMs introduce a transformative approach to KGQA, achieving state-of-the-art results through fine-tuned models and advanced prompting strategies. Earlier works leverage Flan-T5-xl~\cite{chung2022scalinginstructionfinetunedlanguagemodels}, Alpaca-7B~\cite{alpaca}, and LLaMa2-Chat-7B~\cite{touvron2023llama2openfoundation} with CoT (Chain-of-Thought)~\cite{wei2023chainofthoughtpromptingelicitsreasoning}, ToT (Tree-of-Thought)~\cite{yao2023treethoughtsdeliberateproblem}, and GoT (Graph-of-Thought)~\cite{Besta_2024} to enhance reasoning reliability, showcasing strong semantic understanding and reasoning abilities.
Advanced methods highlight these capabilities for KGQA but lead to resource-intensive workflows that lack scalability.
StructGPT~\cite{jiang2023structgptgeneralframeworklarge} unifies the discrepancies between queries and various data formats for collaborative reasoning. ReACT~\cite{yao2023reactsynergizingreasoningacting}, KD-CoT~\cite{wang2023knowledgedrivencotexploringfaithful}, DoG~\cite{ma2025debate}, and ToG~\cite{sun2023thinkongraph} iteratively decompose query and perform step-by-step reasoning while focusing on single-hop subgraphs. However, these methods face challenges such as complex workflows with iterative interactions and increased expenses. KG-CoT~\cite{inproceedings} and RoG~\cite{luo2024reasoninggraphsfaithfulinterpretable} plan relational paths based on the query and extract relevant relational subgraphs from vanilla KGs as external knowledge to assist LLMs, but they are limited to specific KGs and incur high training costs.

Most LLM-based KGQA methods attempt to align query-graph semantics through reasoning, but still face a semantic gap with vanilla KGs, hindering reasoning.


\section{Conclusion}
In this paper, we propose EoG, which leverages LLMs as priors to generate query-aligned graphs for efficient reasoning. For KGQA tasks, we identify the key insight of the semantic gap between queries and graphs and tackle the limitations of vanilla KGs while avoiding the complex reasoning pipeline of existing methods. We also introduce three graph evaluation metrics with theoretical support. Extensive experiments show EoG achieves SOTA performance in KGQA while maintaining low computational costs, scalability, and adaptability across different methods.
\section*{Limitations}
The limitations of our work at the current stage are mainly twofold:

(1) In the Feature Enrich part, we generate a general-domain ontology hierarchy. However, for tasks with varying ontology granularity across different domains, it may be necessary to design domain-specific ontology hierachy structures tailored to each domain.

(2) In our work, we use ChatGPT-series LLMs for query-graph alignment. Further exploration is needed to evaluate the alignment performance of LLMs with different parameter scales. In future work, we will investigate the performance of various LLMs at different parameter levels.


\section{Acknowledgement}

This work is founded by National Natural Science Foundation of China (NSFC62306276 / NSFCU23B2055 / NSFCU19B2027), Zhejiang Provincial Natural Science Foundation of China (No. LQ23F020017), Yongjiang Talent Introduction Programme (2022A-238-G), and Fundamental Research Funds for the Central Universities (226-2023-00138). This work was supported by Ant Group.

\bibliography{custom}

@article{LiuZZFFZHZD26,
  author       = {Yao{-}Yang Liu and
                  Zhen Zheng and
                  Feng Zhang and
                  Jin{-}Cheng Feng and
                  Yi{-}Yang Fu and
                  Jidong Zhai and
                  Bingsheng He and
                  Xiao Zhang and
                  Xiaoyong Du},
  title        = {A comprehensive taxonomy of prompt engineering techniques for large
                  language models},
  journal      = {Frontiers Comput. Sci.},
  volume       = {20},
  number       = {3},
  pages        = {2003601},
  year         = {2026}
}

@inproceedings{ye-etal-2026-tableqa,
    title = "When {T}able{QA} Meets Noise: A Dual Denoising Framework for Complex Questions and Large-scale Tables",
    author = "Ye, Shenghao  and
      Guo, Yu  and
      Jin, Dong  and
      Wang, Yuxiang  and
      Shen, Yikai  and
      Hou, Yunpeng  and
      Chen, Shuangwu  and
      Jianyang  and
      Jiang, Xiaofeng",
    editor = "Liakata, Maria  and
      Moreira, Viviane P.  and
      Zhang, Jiajun  and
      Jurgens, David",
    booktitle = "Proceedings of the 64th Annual Meeting of the {A}ssociation for {C}omputational {L}inguistics (Volume 1: Long Papers)",
    month = jul,
    year = "2026",
    address = "San Diego, California, United States",
    publisher = "Association for Computational Linguistics",
    url = "https://aclanthology.org/2026.acl-long.1102/",
    doi = "10.18653/v1/2026.acl-long.1102",
    pages = "24022--24045",
    ISBN = "979-8-89176-390-6"
}

@inproceedings{guo-etal-2026-rethinking-table,
    title = "Rethinking Table Pruning in {T}able{QA}: From Sequential Revisions to Gold Trajectory-Supervised Parallel Search",
    author = "Guo, Yu  and
      Ye, Shenghao  and
      Chen, Shuangwu  and
      Wen, Zijian  and
      Zhang, Tao  and
      Qirui, Bai  and
      Jin, Dong  and
      Hou, Yunpeng  and
      He, Huasen  and
      Jianyang  and
      Tan, Xiaobin",
    editor = "Liakata, Maria  and
      Moreira, Viviane P.  and
      Zhang, Jiajun  and
      Jurgens, David",
    booktitle = "Proceedings of the 64th Annual Meeting of the {A}ssociation for {C}omputational {L}inguistics (Volume 1: Long Papers)",
    month = jul,
    year = "2026",
    address = "San Diego, California, United States",
    publisher = "Association for Computational Linguistics",
    url = "https://aclanthology.org/2026.acl-long.591/",
    doi = "10.18653/v1/2026.acl-long.591",
    pages = "12960--12976",
    ISBN = "979-8-89176-390-6"
}

@misc{ye2026rethinkingstepwisemodelrouting,
      title={Rethinking Stepwise Model Routing: A Cost-Efficient Table Reasoning Perspective}, 
      author={Shenghao Ye and Yuxiang Wang and Yu Guo and Dong Jin and Shuangwu Chen and Jian Yang},
      year={2026},
      eprint={2605.29319},
      archivePrefix={arXiv},
      primaryClass={cs.CL},
      url={https://arxiv.org/abs/2605.29319}, 
}

@inproceedings{guo-etal-2025-sqlforge,
    title = "{SQLF}orge: Synthesizing Reliable and Diverse Data to Enhance Text-to-{SQL} Reasoning in {LLM}s",
    author = "Guo, Yu  and
      Jin, Dong  and
      Ye, Shenghao  and
      Chen, Shuangwu  and
      Yang, Jian  and
      Tan, Xiaobin",
    editor = "Che, Wanxiang  and
      Nabende, Joyce  and
      Shutova, Ekaterina  and
      Pilehvar, Mohammad Taher",
    booktitle = "Findings of the Association for Computational Linguistics: ACL 2025",
    month = jul,
    year = "2025",
    address = "Vienna, Austria",
    publisher = "Association for Computational Linguistics",
    url = "https://aclanthology.org/2025.findings-acl.443/",
    doi = "10.18653/v1/2025.findings-acl.443",
    pages = "8441--8452",
    ISBN = "979-8-89176-256-5"
}

@inproceedings{miller-etal-2016-key,
    title = "Key-Value Memory Networks for Directly Reading Documents",
    author = "Miller, Alexander  and
      Fisch, Adam  and
      Dodge, Jesse  and
      Karimi, Amir-Hossein  and
      Bordes, Antoine  and
      Weston, Jason",
    editor = "Su, Jian  and
      Duh, Kevin  and
      Carreras, Xavier",
    booktitle = "Proceedings of the 2016 Conference on Empirical Methods in Natural Language Processing",
    month = nov,
    year = "2016",
    address = "Austin, Texas",
    publisher = "Association for Computational Linguistics",
    url = "https://aclanthology.org/D16-1147/",
    doi = "10.18653/v1/D16-1147",
    pages = "1400--1409"
}

@article{liang2026oc,
  title={OC-Distill: Ontology-aware Contrastive Learning with Cross-Modal Distillation for ICU Risk Prediction},
  author={Liang, Zhongyuan and Jo, Junhyung and Lee, Hyang-Jung and Kim, Sang Kyu and Chen, Irene Y},
  journal={arXiv preprint arXiv:2604.16878},
  year={2026}
}

@article{liu-etal-2024-lost,
    title = "Lost in the Middle: How Language Models Use Long Contexts",
    author = "Liu, Nelson F.  and
      Lin, Kevin  and
      Hewitt, John  and
      Paranjape, Ashwin  and
      Bevilacqua, Michele  and
      Petroni, Fabio  and
      Liang, Percy",
    journal = "Transactions of the Association for Computational Linguistics",
    volume = "12",
    year = "2024",
    address = "Cambridge, MA",
    publisher = "MIT Press",
    url = "https://aclanthology.org/2024.tacl-1.9/",
    doi = "10.1162/tacl_a_00638",
    pages = "157--173"
}

@inproceedings{liao-etal-2025-skintern,
    title = "{SKI}ntern: Internalizing Symbolic Knowledge for Distilling Better {C}o{T} Capabilities into Small Language Models",
    author = "Liao, Huanxuan  and
      He, Shizhu  and
      Hao, Yupu  and
      Li, Xiang  and
      Zhang, Yuanzhe  and
      Zhao, Jun  and
      Liu, Kang",
    editor = "Rambow, Owen  and
      Wanner, Leo  and
      Apidianaki, Marianna  and
      Al-Khalifa, Hend  and
      Eugenio, Barbara Di  and
      Schockaert, Steven",
    booktitle = "Proceedings of the 31st International Conference on Computational Linguistics",
    month = jan,
    year = "2025",
    address = "Abu Dhabi, UAE",
    publisher = "Association for Computational Linguistics",
    url = "https://aclanthology.org/2025.coling-main.215/",
    pages = "3203--3221"
}

@inproceedings{DBLP:conf/emnlp/SunDZMSC18,
  author       = {Haitian Sun and
                  Bhuwan Dhingra and
                  Manzil Zaheer and
                  Kathryn Mazaitis and
                  Ruslan Salakhutdinov and
                  William W. Cohen},
  editor       = {Ellen Riloff and
                  David Chiang and
                  Julia Hockenmaier and
                  Jun'ichi Tsujii},
  title        = {Open Domain Question Answering Using Early Fusion of Knowledge Bases
                  and Text},
  booktitle    = {Proceedings of the 2018 Conference on Empirical Methods in Natural
                  Language Processing, Brussels, Belgium, October 31 - November 4, 2018},
  pages        = {4231--4242},
  publisher    = {Association for Computational Linguistics},
  year         = {2018},
  url          = {https://doi.org/10.18653/v1/d18-1455},
  doi          = {10.18653/V1/D18-1455},
  bibsource    = {dblp computer science bibliography, https://dblp.org}
}

@inproceedings{sun-etal-2019-pullnet,
    title = "{P}ull{N}et: Open Domain Question Answering with Iterative Retrieval on Knowledge Bases and Text",
    author = "Sun, Haitian  and
      Bedrax-Weiss, Tania  and
      Cohen, William",
    editor = "Inui, Kentaro  and
      Jiang, Jing  and
      Ng, Vincent  and
      Wan, Xiaojun",
    booktitle = "Proceedings of the 2019 Conference on Empirical Methods in Natural Language Processing and the 9th International Joint Conference on Natural Language Processing (EMNLP-IJCNLP)",
    month = nov,
    year = "2019",
    address = "Hong Kong, China",
    publisher = "Association for Computational Linguistics",
    url = "https://aclanthology.org/D19-1242/",
    doi = "10.18653/v1/D19-1242",
    pages = "2380--2390"
}

@inproceedings{saxena-etal-2020-improving,
    title = "Improving Multi-hop Question Answering over Knowledge Graphs using Knowledge Base Embeddings",
    author = "Saxena, Apoorv  and
      Tripathi, Aditay  and
      Talukdar, Partha",
    editor = "Jurafsky, Dan  and
      Chai, Joyce  and
      Schluter, Natalie  and
      Tetreault, Joel",
    booktitle = "Proceedings of the 58th Annual Meeting of the Association for Computational Linguistics",
    month = jul,
    year = "2020",
    address = "Online",
    publisher = "Association for Computational Linguistics",
    url = "https://aclanthology.org/2020.acl-main.412/",
    doi = "10.18653/v1/2020.acl-main.412",
    pages = "4498--4507"
}

@inproceedings{10.1145/3437963.3441753,
author = {He, Gaole and Lan, Yunshi and Jiang, Jing and Zhao, Wayne Xin and Wen, Ji-Rong},
title = {Improving Multi-hop Knowledge Base Question Answering by Learning Intermediate Supervision Signals},
year = {2021},
isbn = {9781450382977},
publisher = {Association for Computing Machinery},
address = {New York, NY, USA},
url = {https://doi.org/10.1145/3437963.3441753},
doi = {10.1145/3437963.3441753},
booktitle = {Proceedings of the 14th ACM International Conference on Web Search and Data Mining},
pages = {553–561},
numpages = {9},
location = {Virtual Event, Israel},
series = {WSDM '21}
}

@inproceedings{shi-etal-2021-transfernet,
    title = "{T}ransfer{N}et: An Effective and Transparent Framework for Multi-hop Question Answering over Relation Graph",
    author = "Shi, Jiaxin  and
      Cao, Shulin  and
      Hou, Lei  and
      Li, Juanzi  and
      Zhang, Hanwang",
    editor = "Moens, Marie-Francine  and
      Huang, Xuanjing  and
      Specia, Lucia  and
      Yih, Scott Wen-tau",
    booktitle = "Proceedings of the 2021 Conference on Empirical Methods in Natural Language Processing",
    month = nov,
    year = "2021",
    address = "Online and Punta Cana, Dominican Republic",
    publisher = "Association for Computational Linguistics",
    url = "https://aclanthology.org/2021.emnlp-main.341/",
    doi = "10.18653/v1/2021.emnlp-main.341",
    pages = "4149--4158"
}

@inproceedings{zhang-etal-2022-subgraph,
    title = "Subgraph Retrieval Enhanced Model for Multi-hop Knowledge Base Question Answering",
    author = "Zhang, Jing  and
      Zhang, Xiaokang  and
      Yu, Jifan  and
      Tang, Jian  and
      Tang, Jie  and
      Li, Cuiping  and
      Chen, Hong",
    editor = "Muresan, Smaranda  and
      Nakov, Preslav  and
      Villavicencio, Aline",
    booktitle = "Proceedings of the 60th Annual Meeting of the Association for Computational Linguistics (Volume 1: Long Papers)",
    month = may,
    year = "2022",
    address = "Dublin, Ireland",
    publisher = "Association for Computational Linguistics",
    url = "https://aclanthology.org/2022.acl-long.396/",
    doi = "10.18653/v1/2022.acl-long.396",
    pages = "5773--5784"
}

@inproceedings{sun2020sparqaskeletonbasedsemanticparsing,
  title={SPARQA: skeleton-based semantic parsing for complex questions over knowledge bases},
  author={Sun, Yawei and Zhang, Lingling and Cheng, Gong and Qu, Yuzhong},
  booktitle={Proceedings of the AAAI conference on artificial intelligence},
  volume={34},
  number={05},
  pages={8952--8959},
  year={2020}
}

@inproceedings{chen-etal-2019-uhop,
    title = "{UH}op: An Unrestricted-Hop Relation Extraction Framework for Knowledge-Based Question Answering",
    author = "Chen, Zi-Yuan  and
      Chang, Chih-Hung  and
      Chen, Yi-Pei  and
      Nayak, Jijnasa  and
      Ku, Lun-Wei",
    editor = "Burstein, Jill  and
      Doran, Christy  and
      Solorio, Thamar",
    booktitle = "Proceedings of the 2019 Conference of the North {A}merican Chapter of the Association for Computational Linguistics: Human Language Technologies, Volume 1 (Long and Short Papers)",
    month = jun,
    year = "2019",
    address = "Minneapolis, Minnesota",
    publisher = "Association for Computational Linguistics",
    url = "https://aclanthology.org/N19-1031/",
    doi = "10.18653/v1/N19-1031",
    pages = "345--356"
}

@inproceedings{ijcai2019p701,
  title     = {Knowledge Base Question Answering with Topic Units},
  author    = {Lan, Yunshi and Wang, Shuohang and Jiang, Jing},
  booktitle = {Proceedings of the Twenty-Eighth International Joint Conference on
               Artificial Intelligence, {IJCAI-19}},
  publisher = {International Joint Conferences on Artificial Intelligence Organization},
  pages     = {5046--5052},
  year      = {2019},
  month     = {7},
  doi       = {10.24963/ijcai.2019/701},
  url       = {https://doi.org/10.24963/ijcai.2019/701},
}

@inproceedings{lan-jiang-2020-query,
    title = "Query Graph Generation for Answering Multi-hop Complex Questions from Knowledge Bases",
    author = "Lan, Yunshi  and
      Jiang, Jing",
    editor = "Jurafsky, Dan  and
      Chai, Joyce  and
      Schluter, Natalie  and
      Tetreault, Joel",
    booktitle = "Proceedings of the 58th Annual Meeting of the Association for Computational Linguistics",
    month = jul,
    year = "2020",
    address = "Online",
    publisher = "Association for Computational Linguistics",
    url = "https://aclanthology.org/2020.acl-main.91/",
    doi = "10.18653/v1/2020.acl-main.91",
    pages = "969--974"
}

@inproceedings{
jiang2023unikgqa,
title={Uni{KGQA}: Unified Retrieval and Reasoning for Solving Multi-hop Question Answering Over Knowledge Graph},
author={Jinhao Jiang and Kun Zhou and Xin Zhao and Ji-Rong Wen},
booktitle={The Eleventh International Conference on Learning Representations },
year={2023},
url={https://openreview.net/forum?id=Z63RvyAZ2Vh}
}

@inproceedings{ye-etal-2022-rng,
    title = "{RNG}-{KBQA}: Generation Augmented Iterative Ranking for Knowledge Base Question Answering",
    author = "Ye, Xi  and
      Yavuz, Semih  and
      Hashimoto, Kazuma  and
      Zhou, Yingbo  and
      Xiong, Caiming",
    editor = "Muresan, Smaranda  and
      Nakov, Preslav  and
      Villavicencio, Aline",
    booktitle = "Proceedings of the 60th Annual Meeting of the Association for Computational Linguistics (Volume 1: Long Papers)",
    month = may,
    year = "2022",
    address = "Dublin, Ireland",
    publisher = "Association for Computational Linguistics",
    url = "https://aclanthology.org/2022.acl-long.417/",
    doi = "10.18653/v1/2022.acl-long.417",
    pages = "6032--6043"
}

@article{chung2022scalinginstructionfinetunedlanguagemodels,
author = {Chung, Hyung Won and Hou, Le and Longpre, Shayne and Zoph, Barret and Tai, Yi and Fedus, William and Li, Yunxuan and Wang, Xuezhi and Dehghani, Mostafa and Brahma, Siddhartha and Webson, Albert and Gu, Shixiang Shane and Dai, Zhuyun and Suzgun, Mirac and Chen, Xinyun and Chowdhery, Aakanksha and Castro-Ros, Alex and Pellat, Marie and Robinson, Kevin and Valter, Dasha and Narang, Sharan and Mishra, Gaurav and Yu, Adams and Zhao, Vincent and Huang, Yanping and Dai, Andrew and Yu, Hongkun and Petrov, Slav and Chi, Ed H. and Dean, Jeff and Devlin, Jacob and Roberts, Adam and Zhou, Denny and Le, Quoc V. and Wei, Jason},
title = {Scaling instruction-finetuned language models},
year = {2024},
issue_date = {January 2024},
publisher = {JMLR.org},
volume = {25},
number = {1},
issn = {1532-4435},
journal = {J. Mach. Learn. Res.},
month = jan,
articleno = {70},
numpages = {53}
}

@misc{alpaca,
  author = {Rohan Taori and Ishaan Gulrajani and Tianyi Zhang and Yann Dubois and Xuechen Li and Carlos Guestrin and Percy Liang and Tatsunori B. Hashimoto },
  title = {Stanford Alpaca: An Instruction-following LLaMA model},
  year = {2023},
  publisher = {GitHub},
  journal = {GitHub repository},
  howpublished = {\url{https://github.com/tatsu-lab/stanford_alpaca}},
}

@article{touvron2023llama2openfoundation,
  title={Llama 2: Open foundation and fine-tuned chat models},
  author={Touvron, Hugo and Martin, Louis and Stone, Kevin and Albert, Peter and Almahairi, Amjad and Babaei, Yasmine and Bashlykov, Nikolay and Batra, Soumya and Bhargava, Prajjwal and Bhosale, Shruti and others},
  journal={arXiv preprint arXiv:2307.09288},
  year={2023}
}

@article{wei2023chainofthoughtpromptingelicitsreasoning,
  title={Chain-of-thought prompting elicits reasoning in large language models},
  author={Wei, Jason and Wang, Xuezhi and Schuurmans, Dale and Bosma, Maarten and Xia, Fei and Chi, Ed and Le, Quoc V and Zhou, Denny and others},
  journal={Advances in Neural Information Processing Systems},
  volume={35},
  pages={24824--24837},
  year={2022}
}

@article{yao2023treethoughtsdeliberateproblem,
  title={Tree of thoughts: Deliberate problem solving with large language models},
  author={Yao, Shunyu and Yu, Dian and Zhao, Jeffrey and Shafran, Izhak and Griffiths, Tom and Cao, Yuan and Narasimhan, Karthik},
  journal={Advances in Neural Information Processing Systems},
  volume={36},
  pages={11809--11822},
  year={2023}
}

@article{Besta_2024,
   title={Graph of Thoughts: Solving Elaborate Problems with Large Language Models},
   volume={38},
   ISSN={2159-5399},
   url={http://dx.doi.org/10.1609/aaai.v38i16.29720},
   DOI={10.1609/aaai.v38i16.29720},
   number={16},
   journal={Proceedings of the AAAI Conference on Artificial Intelligence},
   publisher={Association for the Advancement of Artificial Intelligence (AAAI)},
   author={Besta, Maciej and Blach, Nils and Kubicek, Ales and Gerstenberger, Robert and Podstawski, Michal and Gianinazzi, Lukas and Gajda, Joanna and Lehmann, Tomasz and Niewiadomski, Hubert and Nyczyk, Piotr and Hoefler, Torsten},
   year={2024},
   month=mar, pages={17682–17690} }

@inproceedings{yao2023reactsynergizingreasoningacting,
  title={React: Synergizing reasoning and acting in language models},
  author={Yao, Shunyu and Zhao, Jeffrey and Yu, Dian and Du, Nan and Shafran, Izhak and Narasimhan, Karthik and Cao, Yuan},
  booktitle={International Conference on Learning Representations},
  year={2023}
}

@article{jiang2023structgptgeneralframeworklarge,
  title={Structgpt: A general framework for large language model to reason over structured data},
  author={Jiang, Jinhao and Zhou, Kun and Dong, Zican and Ye, Keming and Zhao, Wayne Xin and Wen, Ji-Rong},
  journal={arXiv preprint arXiv:2305.09645},
  year={2023}
}

@article{wang2023knowledgedrivencotexploringfaithful,
  title={Knowledge-driven cot: Exploring faithful reasoning in llms for knowledge-intensive question answering},
  author={Wang, Keheng and Duan, Feiyu and Wang, Sirui and Li, Peiguang and Xian, Yunsen and Yin, Chuantao and Rong, Wenge and Xiong, Zhang},
  journal={arXiv preprint arXiv:2308.13259},
  year={2023}
}

@inproceedings{inproceedings,
author = {Zhao, Ruilin and Zhao, Feng and Wang, Long and Wang, Xianzhi and Xu, Guandong},
year = {2024},
month = {08},
pages = {6642-6650},
title = {KG-CoT: Chain-of-Thought Prompting of Large Language Models over Knowledge Graphs for Knowledge-Aware Question Answering},
doi = {10.24963/ijcai.2024/734}
}

@inproceedings{luo2024reasoninggraphsfaithfulinterpretable,
title={Reasoning on Graphs: Faithful and Interpretable Large Language Model Reasoning},
author={Luo, Linhao and Li, Yuan-Fang and Haffari, Gholamreza and Pan, Shirui},
booktitle={International Conference on Learning Representations},
  year={2024}
}

@article{sun2023thinkongraph,
  title={Think-on-graph: Deep and responsible reasoning of large language model on knowledge graph},
  author={Sun, Jiashuo and Xu, Chengjin and Tang, Lumingyuan and Wang, Saizhuo and Lin, Chen and Gong, Yeyun and Ni, Lionel M and Shum, Heung-Yeung and Guo, Jian},
  journal={arXiv preprint arXiv:2307.07697},
  year={2023}
}

@inproceedings{ma2025debate,
  title={Debate on graph: a flexible and reliable reasoning framework for large language models},
  author={Ma, Jie and Gao, Zhitao and Chai, Qi and Sun, Wangchun and Wang, Pinghui and Pei, Hongbin and Tao, Jing and Song, Lingyun and Liu, Jun and Zhang, Chen and others},
  booktitle={Proceedings of the AAAI Conference on Artificial Intelligence},
  pages={24768--24776},
  year={2025}
}

@inproceedings{DBLP:conf/aaai/ZhangZMWS0SZ24,
  author       = {Kun Zhang and
                  Jiali Zeng and
                  Fandong Meng and
                  Yuanzhuo Wang and
                  Shiqi Sun and
                  Long Bai and
                  Huawei Shen and
                  Jie Zhou},
  editor       = {Michael J. Wooldridge and
                  Jennifer G. Dy and
                  Sriraam Natarajan},
  title        = {Tree-of-Reasoning Question Decomposition for Complex Question Answering
                  with Large Language Models},
  booktitle    = {Thirty-Eighth {AAAI} Conference on Artificial Intelligence, {AAAI}
                  2024, Thirty-Sixth Conference on Innovative Applications of Artificial
                  Intelligence, {IAAI} 2024, Fourteenth Symposium on Educational Advances
                  in Artificial Intelligence, {EAAI} 2014, February 20-27, 2024, Vancouver,
                  Canada},
  pages        = {19560--19568},
  publisher    = {{AAAI} Press},
  year         = {2024},
  url          = {https://doi.org/10.1609/aaai.v38i17.29928},
  doi          = {10.1609/AAAI.V38I17.29928},
  bibsource    = {dblp computer science bibliography, https://dblp.org}
}

@article{article,
author = {Berners-Lee, Tim and Hendler, James and Lassila, Ora},
year = {2001},
month = {05},
pages = {},
title = {The Semantic Web: A New Form of Web Content That is Meaningful to Computers Will Unleash a Revolution of New Possibilities},
journal = {ScientificAmerican.com}
}

@article{han2025retrievalaugmentedgenerationgraphsgraphrag,
  title={Retrieval-augmented generation with graphs (graphrag)},
  author={Han, Haoyu and Wang, Yu and Shomer, Harry and Guo, Kai and Ding, Jiayuan and Lei, Yongjia and Halappanavar, Mahantesh and Rossi, Ryan A and Mukherjee, Subhabrata and Tang, Xianfeng and others},
  journal={arXiv preprint arXiv:2501.00309},
  year={2024}
}

@article{article2,
  title={OWL web ontology language overview},
  author={McGuinness, Deborah L and Van Harmelen, Frank and others},
  journal={World Wide Web Consortium Recommendation},
  volume={10},
  number={10},
  pages={2004},
  year={2004}
}

@article{panda2024holmeshyperrelationalknowledgegraphs,
  title={HOLMES: Hyper-relational knowledge graphs for multi-hop question answering using LLMs},
  author={Panda, Pranoy and Agarwal, Ankush and Devaguptapu, Chaitanya and Kaul, Manohar and others},
  journal={arXiv preprint arXiv:2406.06027},
  year={2024}
}

@inproceedings{yasunaga2022qagnnreasoninglanguagemodels,
    title = "{QA}-{GNN}: Reasoning with Language Models and Knowledge Graphs for Question Answering",
    author = "Yasunaga, Michihiro  and
      Ren, Hongyu  and
      Bosselut, Antoine  and
      Liang, Percy  and
      Leskovec, Jure",
    editor = "Toutanova, Kristina  and
      Rumshisky, Anna  and
      Zettlemoyer, Luke  and
      Hakkani-Tur, Dilek  and
      Beltagy, Iz  and
      Bethard, Steven  and
      Cotterell, Ryan  and
      Chakraborty, Tanmoy  and
      Zhou, Yichao",
    booktitle = "Proceedings of the 2021 Conference of the North American Chapter of the Association for Computational Linguistics: Human Language Technologies",
    month = jun,
    year = "2021",
    address = "Online",
    publisher = "Association for Computational Linguistics",
    url = "https://aclanthology.org/2021.naacl-main.45/",
    doi = "10.18653/v1/2021.naacl-main.45",
    pages = "535--546"
}

@inproceedings{10.1145/3534540.3534691,
author = {Guan, Mingyu and Iyer, Anand Padmanabha and Kim, Taesoo},
title = {DynaGraph: dynamic graph neural networks at scale},
year = {2022},
isbn = {9781450393843},
publisher = {Association for Computing Machinery},
address = {New York, NY, USA},
url = {https://doi.org/10.1145/3534540.3534691},
doi = {10.1145/3534540.3534691},
booktitle = {Proceedings of the 5th ACM SIGMOD Joint International Workshop on Graph Data Management Experiences \& Systems (GRADES) and Network Data Analytics (NDA)},
articleno = {6},
numpages = {10},
location = {Philadelphia, Pennsylvania},
series = {GRADES-NDA '22}
}

@inproceedings{DBLP:conf/acl/YihRMCS16,
  author       = {Wen{-}tau Yih and
                  Matthew Richardson and
                  Christopher Meek and
                  Ming{-}Wei Chang and
                  Jina Suh},
  title        = {The Value of Semantic Parse Labeling for Knowledge Base Question Answering},
  booktitle    = {Proceedings of the 54th Annual Meeting of the Association for Computational
                  Linguistics, {ACL} 2016, August 7-12, 2016, Berlin, Germany, Volume
                  2: Short Papers},
  publisher    = {The Association for Computer Linguistics},
  year         = {2016},
  url          = {https://doi.org/10.18653/v1/p16-2033},
  doi          = {10.18653/V1/P16-2033},
  bibsource    = {dblp computer science bibliography, https://dblp.org}
}

@article{Talmor2018TheWA,
  title={The Web as a Knowledge-Base for Answering Complex Questions},
  author={Alon Talmor and Jonathan Berant},
  journal={ArXiv},
  year={2018},
  volume={abs/1803.06643},
  url={https://api.semanticscholar.org/CorpusID:3986974}
}

@inproceedings{Bollacker2008FreebaseAC,
  title={Freebase: a collaboratively created graph database for structuring human knowledge},
  author={Kurt D. Bollacker and Colin Evans and Praveen K. Paritosh and Tim Sturge and Jamie Taylor},
  booktitle={SIGMOD Conference},
  year={2008},
  url={https://api.semanticscholar.org/CorpusID:207167677}
}

@article{yao2019kgbertbertknowledgegraph,
  title={KG-BERT: BERT for knowledge graph completion},
  author={Yao, Liang and Mao, Chengsheng and Luo, Yuan},
  journal={arXiv preprint arXiv:1909.03193},
  year={2019}
}

@inproceedings{reimers2019sentencebertsentenceembeddingsusing,
    title = "Sentence-{BERT}: Sentence Embeddings using {S}iamese {BERT}-Networks",
    author = "Reimers, Nils  and
      Gurevych, Iryna",
    editor = "Inui, Kentaro  and
      Jiang, Jing  and
      Ng, Vincent  and
      Wan, Xiaojun",
    booktitle = "Proceedings of the 2019 Conference on Empirical Methods in Natural Language Processing and the 9th International Joint Conference on Natural Language Processing (EMNLP-IJCNLP)",
    month = nov,
    year = "2019",
    address = "Hong Kong, China",
    publisher = "Association for Computational Linguistics",
    url = "https://aclanthology.org/D19-1410/",
    doi = "10.18653/v1/D19-1410",
    pages = "3982--3992"
}

@article{openai2024gpt4technicalreport,
  title={Gpt-4 technical report},
  author={Achiam, Josh and Adler, Steven and Agarwal, Sandhini and Ahmad, Lama and Akkaya, Ilge and Aleman, Florencia Leoni and Almeida, Diogo and Altenschmidt, Janko and Altman, Sam and Anadkat, Shyamal and others},
  journal={arXiv preprint arXiv:2303.08774},
  year={2023}
}

@article{brown2020languagemodelsfewshotlearners,
  title={Language models are few-shot learners},
  author={Brown, Tom and Mann, Benjamin and Ryder, Nick and Subbiah, Melanie and Kaplan, Jared D and Dhariwal, Prafulla and Neelakantan, Arvind and Shyam, Pranav and Sastry, Girish and Askell, Amanda and others},
  journal={Advances in Neural Information Processing Systems},
  volume={33},
  pages={1877--1901},
  year={2020}
}

@article{chowdhery2022palmscalinglanguagemodeling,
  title={Palm: Scaling language modeling with pathways},
  author={Chowdhery, Aakanksha and Narang, Sharan and Devlin, Jacob and Bosma, Maarten and Mishra, Gaurav and Roberts, Adam and Barham, Paul and Chung, Hyung Won and Sutton, Charles and Gehrmann, Sebastian and others},
  journal={Journal of Machine Learning Research},
  volume={24},
  number={240},
  pages={1--113},
  year={2023}
}

@article{khot2023decomposedpromptingmodularapproach,
  title={Decomposed prompting: A modular approach for solving complex tasks},
  author={Khot, Tushar and Trivedi, Harsh and Finlayson, Matthew and Fu, Yao and Richardson, Kyle and Clark, Peter and Sabharwal, Ashish},
  journal={arXiv preprint arXiv:2210.02406},
  year={2022}
}

@article{rawte2023surveyhallucinationlargefoundation,
  title={A survey of hallucination in large foundation models},
  author={Rawte, Vipula and Sheth, Amit and Das, Amitava},
  journal={arXiv preprint arXiv:2309.05922},
  year={2023}
}

@inproceedings{li2023unigenunifiedgenerativeframework,
  title={Unigen: A unified generative framework for retrieval and question answering with large language models},
  author={Li, Xiaoxi and Zhou, Yujia and Dou, Zhicheng},
  booktitle={Proceedings of the AAAI Conference on Artificial Intelligence},
  volume={38},
  number={8},
  pages={8688--8696},
  year={2024}
}

@article{Ji_2023,
   title={Survey of Hallucination in Natural Language Generation},
   volume={55},
   ISSN={1557-7341},
   url={http://dx.doi.org/10.1145/3571730},
   DOI={10.1145/3571730},
   number={12},
   journal={ACM Computing Surveys},
   publisher={Association for Computing Machinery (ACM)},
   author={Ji, Ziwei and Lee, Nayeon and Frieske, Rita and Yu, Tiezheng and Su, Dan and Xu, Yan and Ishii, Etsuko and Bang, Ye Jin and Madotto, Andrea and Fung, Pascale},
   year={2023},
   month=mar, pages={1–38} }

@article{yuan2024hybridragcomprehensiveenhancement,
  title={A Hybrid RAG System with Comprehensive Enhancement on Complex Reasoning},
  author={Yuan, Ye and Liu, Chengwu and Yuan, Jingyang and Sun, Gongbo and Li, Siqi and Zhang, Ming},
  journal={arXiv preprint arXiv:2408.05141},
  year={2024}
}

@inproceedings{DBLP:conf/semweb/AuerBKLCI07,
  author       = {S{\"{o}}ren Auer and
                  Christian Bizer and
                  Georgi Kobilarov and
                  Jens Lehmann and
                  Richard Cyganiak and
                  Zachary G. Ives},
  editor       = {Karl Aberer and
                  Key{-}Sun Choi and
                  Natasha Fridman Noy and
                  Dean Allemang and
                  Kyung{-}Il Lee and
                  Lyndon J. B. Nixon and
                  Jennifer Golbeck and
                  Peter Mika and
                  Diana Maynard and
                  Riichiro Mizoguchi and
                  Guus Schreiber and
                  Philippe Cudr{\'{e}}{-}Mauroux},
  title        = {DBpedia: {A} Nucleus for a Web of Open Data},
  booktitle    = {The Semantic Web, 6th International Semantic Web Conference, 2nd Asian
                  Semantic Web Conference, {ISWC} 2007 + {ASWC} 2007, Busan, Korea,
                  November 11-15, 2007},
  series       = {Lecture Notes in Computer Science},
  volume       = {4825},
  pages        = {722--735},
  publisher    = {Springer},
  year         = {2007},
  url          = {https://doi.org/10.1007/978-3-540-76298-0\_52},
  doi          = {10.1007/978-3-540-76298-0\_52},
  bibsource    = {dblp computer science bibliography, https://dblp.org}
}

@inproceedings{freebase,
author = {Bollacker, Kurt and Evans, Colin and Paritosh, Praveen and Sturge, Tim and Taylor, Jamie},
year = {2008},
month = {06},
pages = {1247-1250},
title = {Freebase: A collaboratively created graph database for structuring human knowledge},
journal = {Proc. Sigmod},
doi = {10.1145/1376616.1376746}
}

@article{chen2024large,
  title={Large Knowledge Model: Perspectives and Challenges},
  author={Chen, Huajun},
  journal={Data Intelligence},
  volume={6},
  number={3},
  pages={587--620},
  year={2024},
  publisher={China Science Publishing & Media Ltd.},
  doi={10.3724/2096-7004.di.2024.0001}
}

@article{azaria2024chat,
  title={ChatGPT is a Remarkable Tool—For Experts},
  author={Azaria, Amos and Azoulay, Rina and Reches, Shulamit},
  journal={Data Intelligence},
  volume={6},
  number={1},
  pages={240--296},
  year={2024},
  publisher={China Science Publishing & Media Ltd.},
  doi={10.1162/dint_a_00235}
}

@article{zhao2024generating,
  title={AWeCita: Generating Answer with Appropriate and Well-grained Citations Using LLMs},
  author={Zhao, Suifeng and Zhou, Tong and Jin, Zhuoran and Yuan, Hongbang and Chen, Yubo and Liu, Kang and Li, Sujian},
  journal={Data Intelligence},
  volume={6},
  number={4},
  pages={1134--1157},
  year={2024},
  publisher={China Science Publishing & Media Ltd.},
  doi={10.3724/2096-7004.di.2024.0035}
}

@article{li2023evaluation,
  title={Evaluation on ChatGPT for Chinese Language Understanding},
  author={Li, Linhan and Zhang, Huaping and Li, Chunjin and You, Haowen and Cui, Wenyao},
  journal={Data Intelligence},
  volume={5},
  number={4},
  pages={885--903},
  year={2023},
  publisher={China Science Publishing & Media Ltd.},
  doi={10.1162/dint_a_00232}
}

@inproceedings{10.1145/3442381.3449992,
author = {Gu, Yu and Kase, Sue and Vanni, Michelle and Sadler, Brian and Liang, Percy and Yan, Xifeng and Su, Yu},
title = {Beyond I.I.D.: Three Levels of Generalization for Question Answering on Knowledge Bases},
year = {2021},
isbn = {9781450383127},
publisher = {Association for Computing Machinery},
address = {New York, NY, USA},
url = {https://doi.org/10.1145/3442381.3449992},
doi = {10.1145/3442381.3449992},
booktitle = {Proceedings of the Web Conference 2021},
pages = {3477–3488},
numpages = {12},
location = {Ljubljana, Slovenia},
series = {WWW '21}
}

@misc{perevalov2022qald9plusmultilingualdatasetquestion,
      title={QALD-9-plus: A Multilingual Dataset for Question Answering over DBpedia and Wikidata Translated by Native Speakers}, 
      author={Aleksandr Perevalov and Dennis Diefenbach and Ricardo Usbeck and Andreas Both},
      year={2022},
      eprint={2202.00120},
      archivePrefix={arXiv},
      primaryClass={cs.CL},
      url={https://arxiv.org/abs/2202.00120}, 
}

@article{10.1145/2629489,
author = {Vrande\v{c}i\'{c}, Denny and Kr\"{o}tzsch, Markus},
title = {Wikidata: a free collaborative knowledgebase},
year = {2014},
issue_date = {October 2014},
publisher = {Association for Computing Machinery},
address = {New York, NY, USA},
volume = {57},
number = {10},
issn = {0001-0782},
url = {https://doi.org/10.1145/2629489},
doi = {10.1145/2629489},
journal = {Commun. ACM},
month = sep,
pages = {78–85},
numpages = {8}
}

@inproceedings{shu-etal-2022-tiara,
    title = "{TIARA}: Multi-grained Retrieval for Robust Question Answering over Large Knowledge Base",
    author = {Shu, Yiheng  and
      Yu, Zhiwei  and
      Li, Yuhan  and
      Karlsson, B{\"o}rje F.  and
      Ma, Tingting  and
      Qu, Yuzhong  and
      Lin, Chin-Yew},
    editor = "Goldberg, Yoav  and
      Kozareva, Zornitsa  and
      Zhang, Yue",
    booktitle = "Proceedings of the 2022 Conference on Empirical Methods in Natural Language Processing",
    month = dec,
    year = "2022",
    address = "Abu Dhabi, United Arab Emirates",
    publisher = "Association for Computational Linguistics",
    url = "https://aclanthology.org/2022.emnlp-main.555/",
    doi = "10.18653/v1/2022.emnlp-main.555",
    pages = "8108--8121"
}

@misc{yu2023decafjointdecodinganswers,
      title={DecAF: Joint Decoding of Answers and Logical Forms for Question Answering over Knowledge Bases}, 
      author={Donghan Yu and Sheng Zhang and Patrick Ng and Henghui Zhu and Alexander Hanbo Li and Jun Wang and Yiqun Hu and William Wang and Zhiguo Wang and Bing Xiang},
      year={2023},
      eprint={2210.00063},
      archivePrefix={arXiv},
      primaryClass={cs.CL},
      url={https://arxiv.org/abs/2210.00063}, 
}

\section*{Appendix}
\label{sec:appendix}
\appendix
\section{Datasets Statistics}
\label{datastes statics}

We use two KGQA benchmarks that leverage Freebase~\cite{Bollacker2008FreebaseAC} as the underlying knowledge graph: WebQuestionSP~\cite{DBLP:conf/acl/YihRMCS16} and Complex WebQuestions~\cite{Talmor2018TheWA}. To ensure fair comparison, the dataset preprocessing follows previous work~\cite{luo2024reasoninggraphsfaithfulinterpretable}. Dataset statistics are provided in Tab.~\ref{Datasets statistics of WebQSP and CWQ}, which shows the dataset splits as well as the number of reasoning hops required to answer the questions. WebQSP contains 2,826 and 1,628 questions for the training and test sets, respectively, while CWQ consists of 27,639 and 3,531 questions for the training and test sets, respectively. In WebQSP, the questions mainly involve 1-hop and 2-hop reasoning, with 2-hop questions accounting for 34.51\%. CWQ includes questions requiring 1 to 4 hops, with those involving more than 2 hops making up 59.09\%. As shown in Tab.~\ref{Answer Distribution Proportion of WebQSP and CWQ}, both datasets contain multi-answer samples, with 48.8\% of WebQSP questions and 29.4\% of CWQ questions having two or more answers.

\begin{table}[H]
\resizebox{\linewidth}{!}{
\begin{tabular}{c|ccccccc}
\midrule
Datasets & KB & \#Train & \#Test & 1 hop & 2 hop & $\geq 3$ hop & Max \#hop \\ \midrule
WebQSP & Freebase & 2826 & 1628 & 65.49\% & 34.51\% & 0.00\% & 2 \\
CWQ & Freebase & 27639 & 3531 & 40.91\% & 38.34\% & 20.75\% & 4 \\ \midrule
\end{tabular}
}
\caption{Datasets statistics of WebQSP and CWQ}
\label{Datasets statistics of WebQSP and CWQ}
\end{table}

\begin{table}[H]
\centering
\resizebox{0.8\linewidth}{!}{
\begin{tabular}{c|cccc}
\midrule
Datasets & \#Ans=1 & 2 $\geq$ \#Ans $\leq$ & 5 $\geq$ \#Ans $\leq$ 9 & \#Ans $\geq$ 10 \\ \midrule
WebQSP & 51.20\% & 27.40\% & 8.30\% & 12.10\% \\
CWQ & 70.60\% & 19.40\% & 6\% & 4\% \\ \midrule
\end{tabular}
}
\caption{Answer distribution proportion of WebQSP and CWQ}
\label{Answer Distribution Proportion of WebQSP and CWQ}
\end{table}

\section{Prompts}
Listing~\ref{CoT Prompt},~\ref{Query Structuring Prompt},~\ref{Structural Enrich Prompt},~\ref{Feature Enrich Prompt},~\ref{Question Answering Prompt} detail the specific designs of the Chain of Thought prompt, Query Structuring prompt, Structural Enrich prompt, Feature Enrich prompt, and Question Answering prompt. For the CoT prompt, we directly adopt the design from~\cite{sun2023thinkongraph}. Due to space constraints, some demonstrations for the Structural Enrich prompt and Feature Enrich prompt are omitted.
\subsection{Chain of Thought Prompt}

\lstset{
    basicstyle=\ttfamily\fontsize{7}{8}\selectfont,
    breaklines=true,                  
    prebreak=\mbox{},               
    postbreak=\mbox{},              
    breakindent=0pt,               
    frame=single,                  
    tabsize=4,                      
    showstringspaces=false,   
    columns=fixed,           
    keepspaces=true,             
    captionpos=b,
}

\begin{lstlisting}[language=, caption={Demonstration of CoT Prompt}]
Q: What state is home to the university that is represented in sports by George Washington Colonials men's basketball?\nA: First, the education institution has a sports team named George Washington Colonials men's basketball in is George Washington University , Second, George Washington University is in Washington D.C. The answer is {Washington, D.C.}.

Q: Who lists Pramatha Chaudhuri as an influence and wrote Jana Gana Mana?
A: First, Bharoto Bhagyo Bidhata wrote Jana Gana Mana. Second, Bharoto Bhagyo Bidhata lists Pramatha Chaudhuri as an influence. The answer is {Bharoto Bhagyo Bidhata}.

Q: Who was the artist nominated for an award for You Drive Me Crazy?
A: First, the artist nominated for an award for You Drive Me Crazy is Britney Spears. The answer is {Jason Allen Alexander}.

Q: What person born in Siegen influenced the work of Vincent Van Gogh?
A: First, Peter Paul Rubens, Claude Monet and etc. influenced the work of Vincent Van Gogh. Second, Peter Paul Rubens born in Siegen. The answer is {Peter Paul Rubens}.

Q: What is the country close to Russia where Mikheil Saakashvii holds a government position?
A: First, China, Norway, Finland, Estonia and Georgia is close to Russia. Second, Mikheil Saakashvii holds a government position at Georgia. The answer is {Georgia}.

Q: What drug did the actor who portrayed the character Urethane Wheels Guy overdosed on?
A: First, Mitchell Lee Hedberg portrayed character Urethane Wheels Guy. Second, Mitchell Lee Hedberg overdose Heroin. The answer is {Heroin}.

Q: {question}
\end{lstlisting}
\label{CoT Prompt}


\subsection{Query Structuring Prompt}
\begin{lstlisting}[language=, caption={Demonstration of Query Structuring Prompt}]
[INST] <<SYS>>
<</SYS>>
{instruction}
Given a question, decompose it step by step into smaller components until it is broken down into unit queries that can be directly answered without further reasoning or decomposition. A unit query is a question that cannot be divided further and can be answered directly. If the given question is already a unit query, no decomposition is needed(If a question can still be further broken down, it must be divided into at least $two or more sub-questions$. Otherwise, the question is considered a unit query.). Your output should only include the tree structure of the decomposed question, with sub-questions indented using "-", and no additional content should be provided, just the tree structure. The format for the decomposition tree is as follows:
### decomposition tree format ###
question
-sub-question
--sub-sub-question
---sub-sub-sub-question
---sub-sub-sub-question
--sub-sub-question
-sub-question
### decomposition tree format ###
I will provide examples, please complete your task after reviewing them.
{/instruction}
{demonstrations}
### Example 1:
Input:
What is the name of the scientist who developed the theory that explains why objects fall to Earth?
Output:
What is the name of the scientist who developed the theory that explains why objects fall to Earth?
-What is the theory that explains why objects fall to Earth?
--Is there a theory for why objects fall to Earth?
--What is the name of this theory?
-Who developed this theory?
--Is this theory associated with a specific scientist?
--What is the name of this scientist?
### Example 2:
Input:
What type of energy powers the device invented by Thomas Edison that produces light?
Output:
What type of energy powers the device invented by Thomas Edison that produces light?
-What device produces light and was invented by Thomas Edison?
--Who is Thomas Edison?
--What devices did Thomas Edison invent?
--Is there a device invented by Thomas Edison that produces light?
--What is the name of this device?
-What type of energy powers this device?
--What is energy in this context?
--What is the primary mechanism or process that allows this device to produce light?
--What type of energy drives this mechanism?
### Example 3:
Input:
What inspired the author of the book "1984" to write it?
Output:
What inspired the author of the book "1984" to write it?
-Who is the author of the book "1984"?
--What is the book "1984"?
--Who wrote the book "1984"?
-What inspired this author to write "1984"?
--What was happening during the time this author wrote "1984"?
--What personal experiences influenced the author?
--What political or social events might have inspired the author?
--Did any other books or ideas inspire the author?
### Example 4:
Input:
What is the atomic number of the element discovered in the laboratory where the youngest Nobel Physics laureate worked?
Output:
What is the atomic number of the element discovered in the laboratory where the youngest Nobel Physics laureate worked?
-Who is the youngest Nobel Physics laureate?
--When did this person win the Nobel Prize?
--What was their age at the time of the award?
-Where did this laureate work?
--Did this laureate work in laboratory/institution?
--what's the name of the laboratory/institution?
-Was any chemical element discovered at this laboratory?
--What is the name of the element?
--What is its atomic number?
### Example 5:
Input:
Which country hosted the sporting event where the first female gold medalist in track and field competed?
Output:
Which country hosted the sporting event where the first female gold medalist in track and field competed?
-What type of sporting event is being referred to (e.g., Olympics)?
-Who was the first female gold medalist in track and field?
--What is track and field?
--Which female athlete won the first gold medal in this category?
-In which event within track and field did she win?
-Which country hosted this sporting event?
{/demonstrations}
### Your Turn
Input:
{question}
[/INST]
\end{lstlisting}
\label{Query Structuring Prompt}

\subsection{Structural Enrich Prompt}
\begin{lstlisting}[language=, caption={Demonstration of Structural Enrich Prompt}]
[INST] <<SYS>>You are a linguist capable of understanding user queries and the semantic information contained in triples. You can enhance the semantic information of triples based on the principles of similarity, symmetry, and transitivity, bridging the semantic gap between the triples and the user queries. Your goal is to augment the triples semantically so they can better address the user's queries.
<</SYS>>
{instruction}
You are given a subgraph that consists of multiple triples, where each triple corresponds to one or more user queries. The format is as follows:
triple(entity1, relation, entity2) - query1 - query2 - query3 - ...
Here, a query represents a question from the user. The triple associated with a query may contain information relevant to answering the question. However, the semantic connections both within and between triples are typically weak, lacking dense links to clearly establish relationships between entities. As a result, answers to the queries are often implicit and require complex reasoning. To bridge the semantic gap between the queries and the triples, please perform semantic enhancement on the subgraph by considering the user queries.
Specifically, you will utilize the following properties for semantic enhancement: Similarity, Symmetry, and Transitivity. Below, you will find detailed explanations of these properties along with examples. Please keep these principles in mind as you perform semantic enhancement.
{Similarity}
The property of similarity indicates that for a given triple (e1, r1, e2), we can find an alternative relationship r2 that is semantically different from r1 but shares the same directional connection between e1 and e2. In other words, the entities e1 and e2 remain connected in the same direction, but the meaning of the connection changes to reflect the semantics of r2. For example:
If (e1, r1, e2) exists, then (e1, r2, e2) may also exist, where r1 != r2, and r1 and r2 represent different relationships.
{/Similarity}
{Symmetry}
The property of symmetry indicates that for a given triple (e1, r1, e2), there exists a new relationship r2 that is semantically different from r1 and reverses the direction of the connection between e1 and e2. In other words, the new triple swaps the roles of e1 and e2 while preserving their semantic relationship in a reversed form. For example:
If (e1, r1, e2) exists, then (e2, r2, e1) may also exist, where r1 != r2, and r1 and r2 represent different relationships.
{/Symmetry}
{Transitivity}
The property of transitivity indicates that when two triples share a common entity, we can combine their relationships into a new relationship that connects the remaining two entities directly. Specifically, if (e1, r1, e2) and (e2, r2, e3) exist, then there exists a new relationship r3 that semantically encompasses both r1 and r2, forming a new triple (e1, r3, e3). The new relationship r3 captures the combined semantics of r1 and r2. For example:
Given (e1, r1, e2) and (e2, r2, e3), we derive (e1, r3, e3), where r3 combines the meanings of r1 and r2.
{/Transitivity}
Your task will follow the four steps below:
{Procedure}
Step 1:
Examine the given triples and their associated queries. Understand the information contained within the triples and the questions being asked in the queries.
Step 2:
Using the similarity and symmetry properties, perform semantic enhancement on the given 1-hop subgraph (i.e., triples where connections involve only directly linked entities).
Step 3:
Using the transitivity property, perform semantic enhancement on the given multi-hop subgraph (i.e., triples where connections involve intermediary entities).
Step 4:
Output the newly generated triples resulting from the semantic enhancement process.
{/Procedure}
I will provide examples, please complete your task after reviewing them.
{/instruction}
{demonstrations}
### Example 1:
Input:
(Michelle Bachelet,people.person.nationality,Chile)-What is the location that appointed Michelle Bachelet to a governmental position?-Who is Michelle Bachelet?-What governmental position was Michelle Bachelet appointed to?-Where was Michelle Bachelet appointed to this position?
(Chile,language.human_language.countries_spoken_in,Spanish Language)-What language is spoken in this location?
1-hop:
(Michelle Bachelet,people.person.nationality,Chile)
2-hop:
(Michelle Bachelet,people.person.nationality,Chile)->(Chile,language.human_language.countries_spoken_in,Spanish Language)
Output:
{thought}
step 1:
To prepare for the subsequent semantic enhancement of the triples and bridge the semantic gap between the triples and the user queries, let me review the triples along with their corresponding user query(ies):
(Michelle Bachelet,people.person.nationality,Chile)-What is the location that appointed Michelle Bachelet to a governmental position?-Who is Michelle Bachelet?-What governmental position was Michelle Bachelet appointed to?-Where was Michelle Bachelet appointed to this position?
(Chile,language.human_language.countries_spoken_in,Spanish Language)-What language is spoken in this location?
step 2:
Using the similarity property, perform semantic enhancement on the given 1-hop subgraph:
For the 1-hop subgraph (Michelle Bachelet,people.person.nationality,Chile),the related queries are "What is the location that appointed Michelle Bachelet to a governmental position?" , "Who is Michelle Bachelet?" , "What governmental position was Michelle Bachelet appointed to?" , "Where was Michelle Bachelet appointed to this position?" .Combining these queries, The newly added  triple(s) is/are (Michelle Bachelet,appointed_government_position_in,Chile)
Using the symmetry properties, perform semantic enhancement on the given 1-hop subgraph:
For the 1-hop subgraph (Michelle Bachelet,people.person.nationality,Chile),the related queries are "What is the location that appointed Michelle Bachelet to a governmental position?" , "Who is Michelle Bachelet?" , "What governmental position was Michelle Bachelet appointed to?" , "Where was Michelle Bachelet appointed to this position?" .Combining these queries, The newly added  triple(s) is/are (Chile, appointed_as_government_official_by, Michelle Bachelet)
step 3:
Using the transitivity properties, perform semantic enhancement on the given 2-hop subgraph:
For the 2-hop subgraph (Michelle Bachelet,people.person.nationality,Chile)->(Chile,language.human_language.countries_spoken_in,Spanish Language),the related queries are "What is the location that appointed Michelle Bachelet to a governmental position?" , "Who is Michelle Bachelet?" , "What governmental position was Michelle Bachelet appointed to?" , "Where was Michelle Bachelet appointed to this position?" , What language is spoken in this location?. Combining these queries, The newly added  triple(s) is/are (Michelle Bachelet, language_of_the_country_where_appointed ,Spanish Language)
step 4:
Final output:
(Michelle Bachelet,appointed_government_position_in,Chile)
(Chile, appointed_as_government_official_by, Michelle Bachelet)
(Michelle Bachelet, language_of_the_country_where_appointed ,Spanish Language)
{/thought}
...
{/demonstrations}
### Your Turn
Input:
{quadruples}
1-hop:
{1-hop path}
2-hop:
{2-hop path}
[/INST]
\end{lstlisting}
\label{Structural Enrich Prompt}

\subsection{Feature Enrich Prompt}
\begin{lstlisting}[language=, caption={Demonstration of Feature Enrich Prompt}]
[INST] <<SYS>>You are a linguist with extensive expertise in ontology knowledge. You have the ability to understand the context of entities, including user queries and the semantic information of triples. Based on the contextual information of an entity, you can accurately and appropriately add ontologies to the entity. Your goal is to bridge the semantic gap between entities and user queries by adding relevant ontologies, thereby semantically augmenting the entities to enable them to better address the user's queries.
<</SYS>>
{instruction}
You are provided with an entity list. Each entity in the list is accompanied by its context information, which consists of two parts:
1.Relevant triples associated with the entity.
2.Relevant user queries associated with the entity.
The context format for each entity is as follows:
[$entity$ context]
relavent triple(s):triple1-triple2-triple3-...
relavent user query(ies):query1-query2-query3-...
[/$entity$ context]
Your task is to analyze the context information of each entity, apply your expert knowledge, and assign an appropriate ontology to the entity. The added ontology should:
1.Be semantically consistent with the entity's context information.
2.Avoid any contradictions or irrelevance with the given context.
3.Enhance the entity's ability to better answer the user query(ies) in its context.
Store the ontology assignments in a set called {result}. The {result} set is initially empty. Each ontology assignment should follow the format:
{result}
(entity,ontology_relation,newly added ontology)
...
{/result}
You must choose an appropriate ontology relation from the options provided below.
{ontology relation definition}
HYP: Hypernym relation
Refers to the relation where a broader or more general concept includes or encompasses a more specific or specialized concept.
Hypernym_isA
Is a type of...
Hypernym_locateAt
Is located at...
Hypernym_mannerOf 
A is a specific implementation or way of B. Similar to "isA," but used for verbs. For example, "auction" -> "sale"
IND: Induction relation
Refers to the relation between individual entities and conceptual generalizations derived from a class of entities with common characteristics. 
Induction_belongTo
This relation is commonly used in SPG to describe the classification relation from entity types to concept types. For example, "company event" -> "company event category".
INC: Inclusion relation
Expresses the relation between parts and the whole. 
Inclusion_isPartOf
A is a part of B.
Inclusion_madeOf
A is made up of B. For example, "bottle" -> "plastic". 
Inclusion_derivedFrom
A is derived from or originated from B, used to express composite concepts.
Inclusion_hasContext
A is a word used in the context of B, where B can be a subject area, technical field, or regional dialect. For example, "astern" -> "ship".
{/ontology relation definition}
You can only choose an appropriate {ontology_relation} from {Hypernym_isA, Hypernym_locateAt, Hypernym_mannerOf, Induction_belongTo, Inclusion_isPartOf, Inclusion_madeOf, Inclusion_derivedFrom, Inclusion_hasContext}.
Your task will follow the four steps below:
{Procedure}
step 1:
For the first $entity$ in the entity list, read its context information carefully. Based on the provided context and your expert knowledge, answer the following question:
Does this $entity$ already have an ontology assigned?
If yes, proceed to Step 2.
If no, move to Step 3.
step 2:
Using the entity's context information and your expert knowledge:
Identify a list of appropriate ontologies for the entity.
For each ontology in this list, select the most suitable ontology relation from the {ontology_relation} options.
Format each assignment as (entity, ontology_relation, newly added ontology) and add it to the {result} set.
Remove this $entity$ and its corresponding context information from the entity list.
Output the current {result} set in the following format:
{result}
...
{/result}
Then move to Step 3.
step 3:
Check if the entity list is now empty:
If the list is empty, proceed to Step 4.
If the list is not empty, return to Step 1 and repeat the process for the next $entity$.
step 4:
Once all entities have been processed, output the final {result} set containing all (entity, ontology_relation, newly added ontology) assignments in the following format:
{result}
(entity,ontology_relation,newly added ontology)
...
{/result}
{/Procedure}
I will provide examples, please complete your task after reviewing them.
{/instruction}
{demonstrations}
### Example 1:
Input:
entity List:
[$Michelle Bachelet$ context]
relavent triple(s):(Michelle Bachelet,people.person.nationality,Chile)
relavent user query(ies):What is the location that appointed Michelle Bachelet to a governmental position?-Who is Michelle Bachelet?-What governmental position was Michelle Bachelet appointed to?-Where was Michelle Bachelet appointed to this position?
[/$Michelle Bachelet$ context]
[$Chile$ context]
relavent triple(s):(Michelle Bachelet,people.person.nationality,Chile)-(Chile,language.human_language.countries_spoken_in,Spanish Language)
relavent user query(ies):What is the location that appointed Michelle Bachelet to a governmental position?-Who is Michelle Bachelet?-What governmental position was Michelle Bachelet appointed to?-Where was Michelle Bachelet appointed to this position?-What language is spoken in this location?
[/$Chile$ context]
Output:
{result}
(Michelle Bachelet, Hypernym_isA, Political Figure)
(Michelle Bachelet, Hypernym_isA, President)
(Chile, Hypernym_isA, Country)
(Chile, Hypernym_locateAt, South America)
(Chile, Inclusion_hasContext, Spanish Language)
{/result}
...
{/demonstrations}
### Your Turn
Input:
entity List:
[$entity$ context]
relavent triple(s):
relavent user query(ies):
[/$entity$ context]
[/INST]
\end{lstlisting}
\label{Feature Enrich Prompt}

\subsection{Question Answering Prompt}
\begin{lstlisting}[language=, caption={Demonstration of Question Answering Prompt}]
[INST] <<SYS>>
<</SYS>>
{instruction}
You are given an input question along with additional information to assist in answering the question. The additional information includes a set of relevant triples. The format of a triple is as follows: 
(e1, r, e2)
Your task is to perform step-by-step reasoning based on the provided additional information to arrive at the answer. You should output your process of thinking and reasoning then the final answer. Each answer should be as close to a single entity as possible, rather than a long sentence. Each answer should not be in the form of an ID, such as: "m.0n1v8cy", "m.0jx21d", or similar formats. If the provided information is insufficient to infer the correct answer, please use your internal knowledge to generate a response. Please try to output all possible answers you consider correct. If there is only one answer, directly output that answer. If there are multiple answers, separate them using <SEP>.
### Input Format ###
Input:  
question:
{input question}  
information:
{triple1}
{triple2}
{triple3}
...
### Output Format ###
Output:
{thoughts & reason}
Your process of thinking and reasoning
...
{/thoughts & reason}
Final answer:
{answer}
{/instruction}
### Your Turn
Input:
question:
{question}
information:
{knowledge graph}
[/INST]
\end{lstlisting}
\label{Question Answering Prompt}

\section{Details of our Theoretical Proof}
\label{Details of our Theoretical Proof}
In this section, we provide a detailed derivation of the solution for EoG.

As stated in~\ref{theorem1}, our goal is to find an optimized graph $  G^*  $ by maximizing the expected posterior probability:

\[{G^{*} = \underset{G}{argmax}}{~\mathbb{E}_{P(q,G)}\left\lbrack P\left( M_{\theta},q \middle| G \right) \right\rbrack}\]

It is equivalent to maximizing the mutual information between $  q  $ and $  G  $.
The detailed derivation is as follows.

Since $  P(q)  $ and $  P(G)  $ are constants, we have:

{\small
\[P\left( M_{\theta},q \middle| G \right) \propto {\log\left( \frac{P\left( M_{\theta},q \middle| G \right)}{P(q)} \right)}\]
}

The equivalent transformation of the conditional probability is:

{\small
\[P\left( M_{\theta},q \middle| G \right) = \frac{P\left( M_{\theta},G \middle| q \right)P(q)}{P(G)}\]
}

We can obtain:

{\small
\[P\left( M_{\theta},q \middle| G \right) \propto {\log\left( \frac{P\left( M_{\theta},q \middle| G \right)}{P(q)} \right)}\]
}
{\small

\[{{= \log}\left( \frac{P\left( M_{\theta},G \middle| q \right)}{P(G)} \right)}\]

}

We eliminate the variable $  M_\theta  $ using the marginalization formula:

{\small
\[{\mathit{\log}\left( \frac{P\left( M_{\theta},G \middle| q \right)}{P(G)} \right)} \propto {\mathit{\log}\left( \frac{\int_{M_{\theta}}^{}{P\left( M_{\theta},G \middle| q \right)dM_{\theta}}}{P(G)} \right)}\]
}

We can obtain:

{\small

\[{\mathit{\log}\left( \frac{\int_{M_{\theta}}^{}{P\left( M_{\theta},G \middle| q \right)dM_{\theta}}}{P(G)} \right)} = \mathit{\log}\left( \frac{P\left( G \middle| q \right)}{P(G)} \right)\]

}

{\small

\[= {\log\left( \frac{P\left( {q,G} \right)}{P(q)P(G)} \right)}\]

}
Thus, the expectation of the original posterior probability is equivalent to:

{\small

\[\mathbb{E}_{P(Q,G)}\left\lbrack P\left( M_{\theta},q \middle| G \right) \right\rbrack \propto \mathbb{E}_{P(q,G)}\left\lbrack {\log\left( \frac{P\left( {q,G} \right)}{P(q)P(G)} \right)} \right\rbrack\]

}

Finally we can get:

\[\mathbb{E}_{P(q,G)}\left\lbrack {P\left( M_{\theta},q \middle| G \right)} \right\rbrack \propto MI(q,G)\]

\section{Optimal k-value Exploration}
\label{Optimal k-value Exploration}
\begin{figure}[h!]
    \centering
    \includegraphics[width=\linewidth, keepaspectratio]{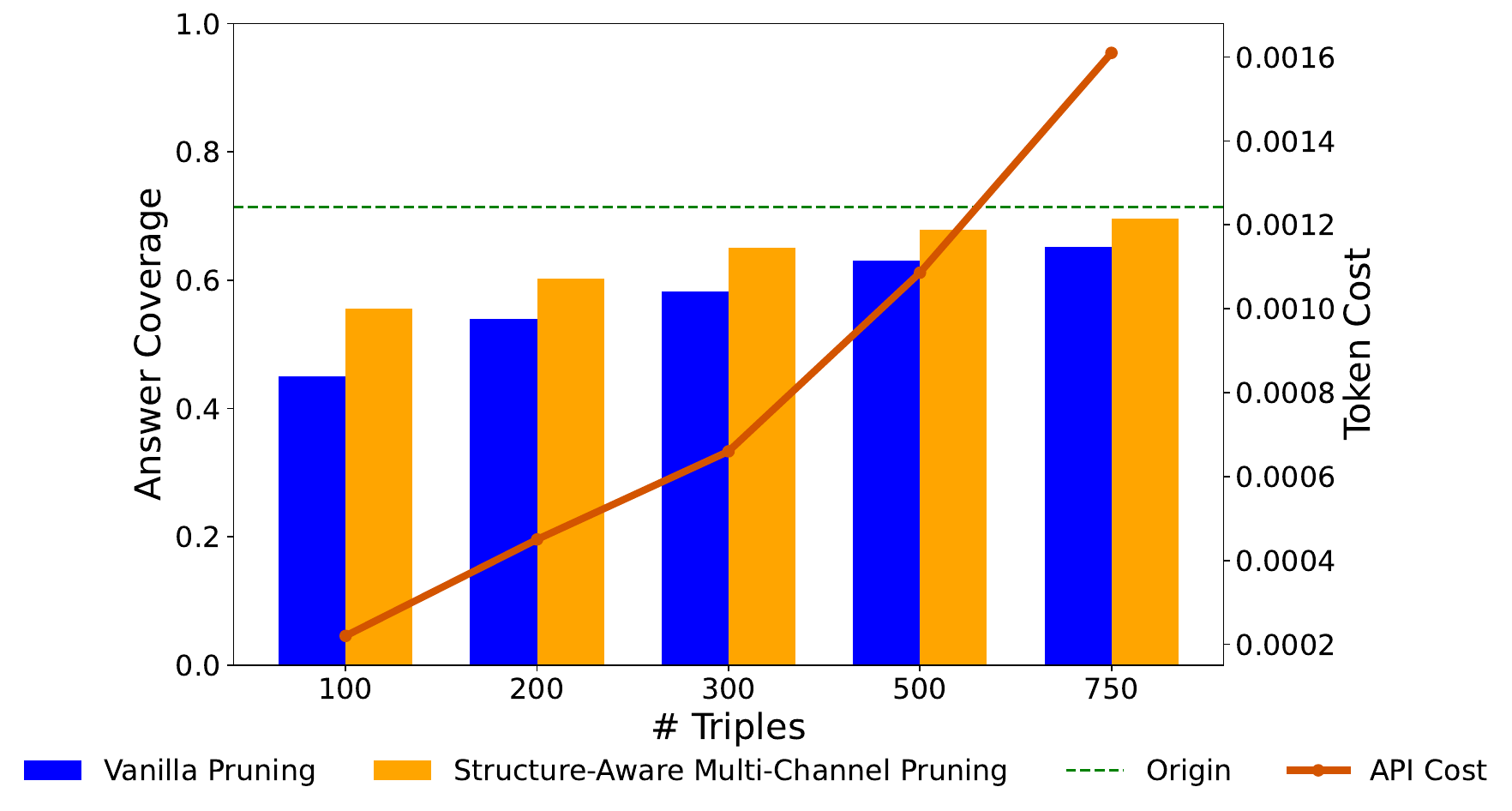}
    \caption{Impact of the Number of Retrieved Triples on Answer Coverage Ratio and API Call Cost (\$) in the CWQ Dataset}
    \label{fig:k cwq}
\end{figure}

\begin{figure}[h!]
    \centering
    \includegraphics[width=\linewidth, keepaspectratio]{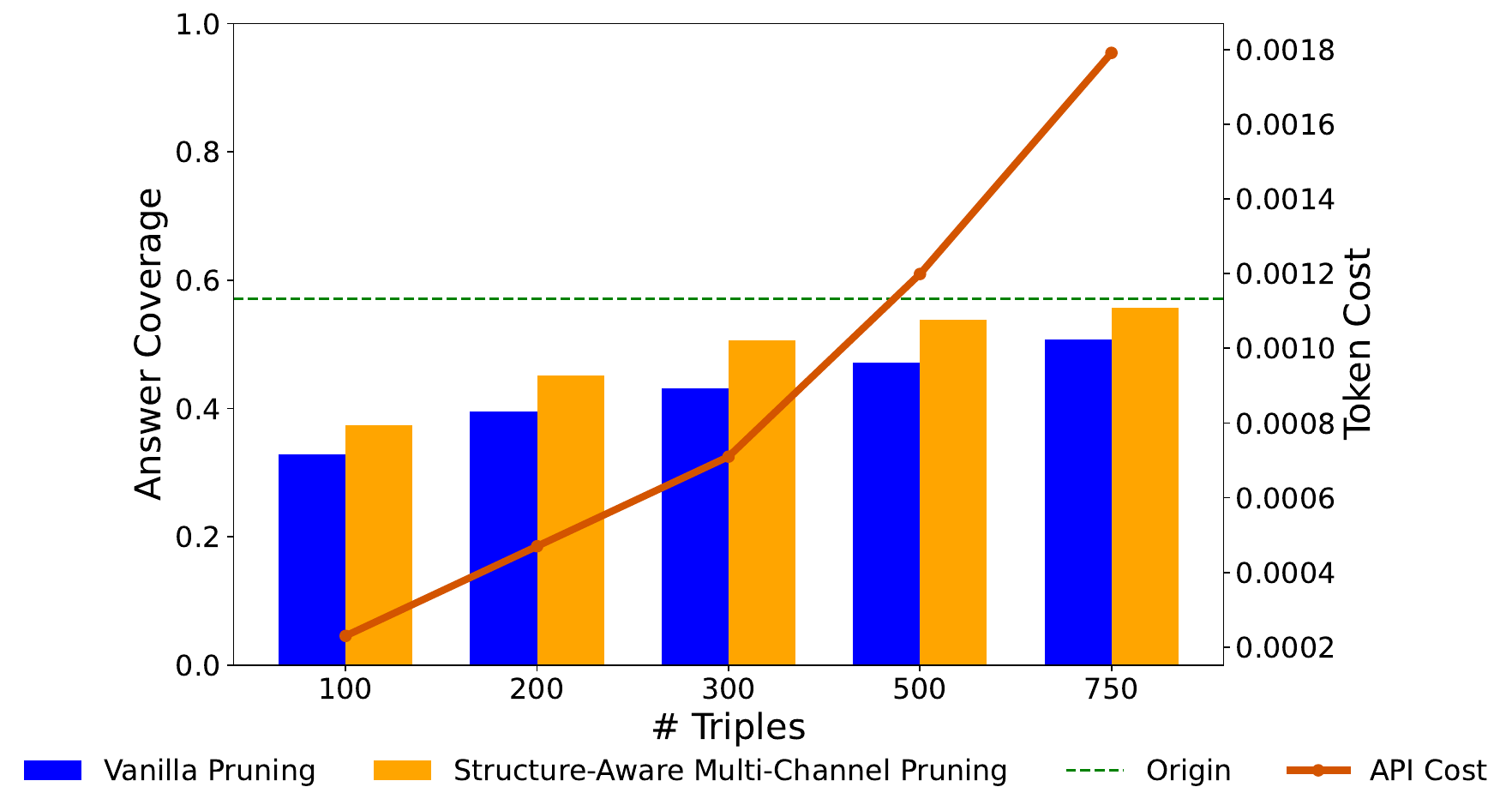}
    \caption{Impact of the Number of Retrieved Triples on Answer Coverage Ratio and API Call Cost (\$) in the WebQSP Dataset}
    \label{fig:k webqsp}
\end{figure}

To explore the optimal number of retrieved triples(\textit{k}) as discussed in Section~\ref{Structure-Aware Multi-Channel Pruning}, and to demonstrate the superiority of our Structure-Aware Multi-Channel Pruning method compared to the vanilla pruning approach, we evaluated the impact of different numbers of retrieved triples on \textbf{Answer Coverage} and \textbf{Token Cost} (using the default GPT-4o-mini) on the WebQSP and CWQ datasets.
For answer coverage, it is calculated as: \# Answers in Pruned Graph / \# Total Answers. For Token Cost, it is determined by converting triples directly into natural language text, e.g., converting (\textit{Beijing, located in, China}) into \textit{Beijing located in China}. We then compute the total token number of the converted text and multiply it by the OpenAI API price to get the token cost. The results are shown in Fig.~\ref{fig:k cwq} and Fig.~\ref{fig:k webqsp}, where the values represented by the green dashed lines indicate the answer coverage before pruning performed by~\cite{luo2024reasoninggraphsfaithfulinterpretable} on the dataset.
As the number of retrieved triples increases, the number of answers mistakenly removed by pruning decreases, but the token length grows, leading to a continuous increase in API invocation costs. When the number of retrieved triples is 300, the pruned graph obtained by Structure-Aware Multi-Channel Pruning has only a 5\% reduction in answer coverage compared to the unpruned graph. Compared to the vanilla pruning method, our pruning strategy achieves approximately 7\% higher answer coverage across all triple counts, which demonstrates the superiority of our approach.

Considering the trade-off between token cost and answer coverage, we chose $k=300$ as the optimal number of retrieved triples.

\section{Case Study}
\label{Case Study}

\begin{figure*}[ht]
    \centering
    \includegraphics[width=\textwidth, keepaspectratio]{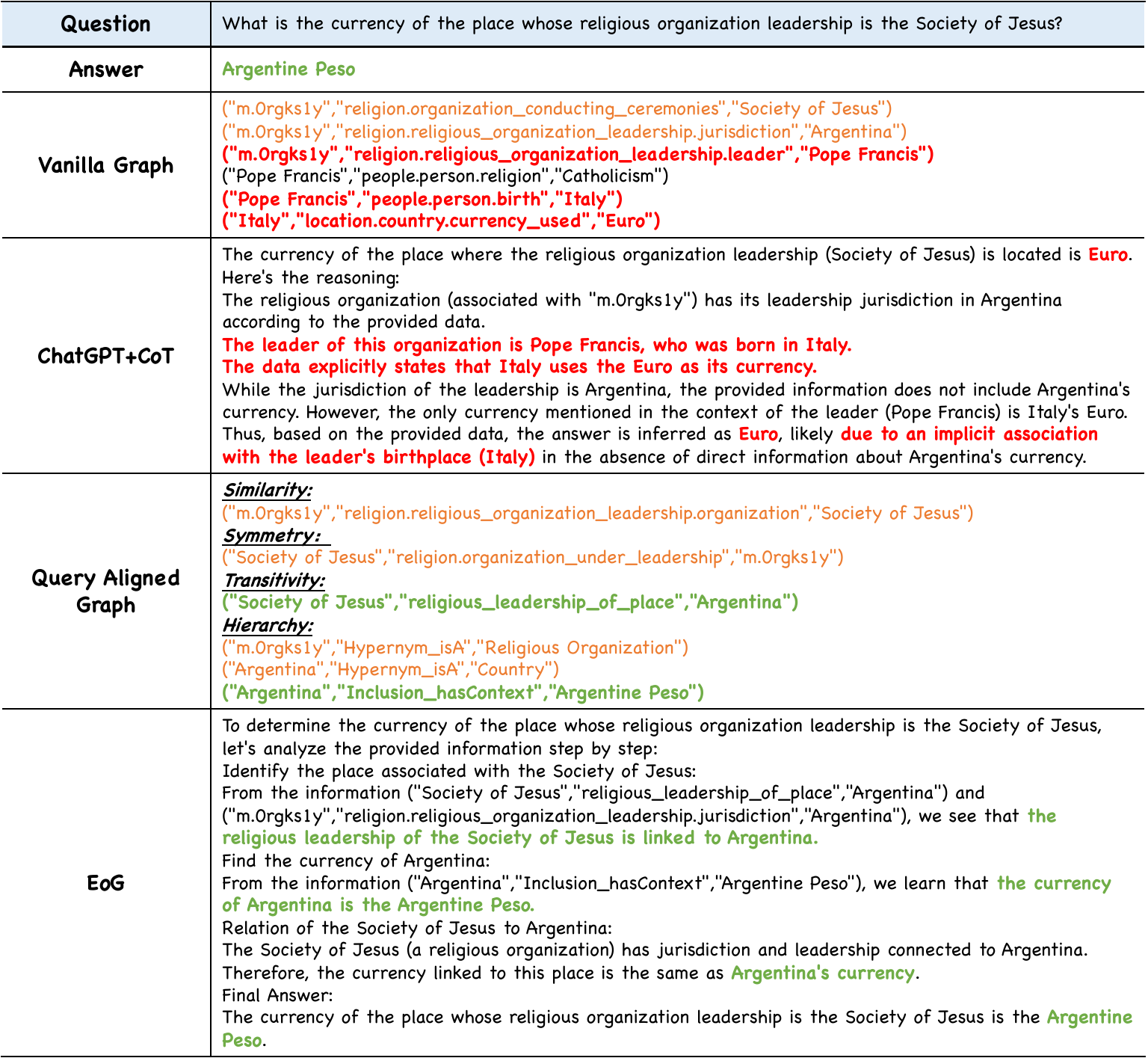}
    \caption{Case for EoG. \textbf{Green} highlights the correct answers or key triples, \textbf{Red} indicates misleading triples, and \textbf{Orange} represents triples relevant to the query.}
    \label{fig:case1}
\end{figure*}

\begin{figure*}[ht]
    \centering
    \includegraphics[width=\textwidth, keepaspectratio]{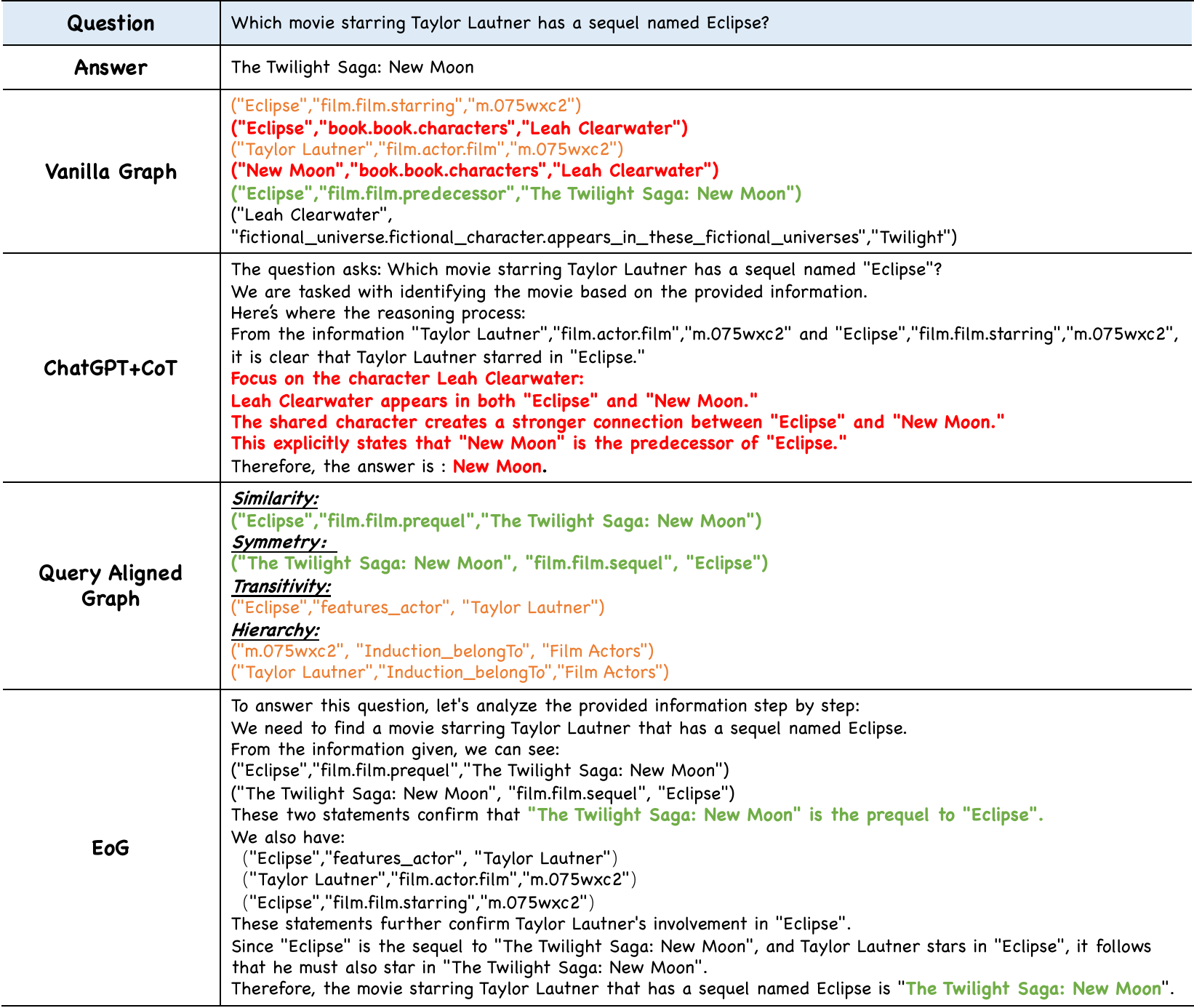}
    \caption{Case for EoG. \textbf{Green} highlights the correct answers or key triples, \textbf{Red} indicates misleading triples, and \textbf{Orange} represents triples relevant to the query.}
    \label{fig:case2}
\end{figure*}

In this section, we present case study on the CWQ dataset to illustrate the critical role of EoG-generated query-aligned graphs in reasoning. We provide two examples, as shown in Fig.~\ref{fig:case1} and Fig.~\ref{fig:case2}, comparing the performance of EoG with the vanilla KG method that assists ChatGPT in step-by-step reasoning. In the figures, green highlights the correct answers or key triples, red indicates misleading triples, and orange represents triples relevant to the query.

In the first example, the query is \textit{What is the currency of the place whose religious organization leadership is the Society of Jesus?} The retrieved triples include misleading information: (\textit{Pope Francis}, \textit{people.person.birth}, \textit{Italy}), (\textit{Italy}, \textit{location.country.currency\_used}, \textit{Euro}).
Using the vanilla KG, ChatGPT retrieves the entity \textit{m.0rgks1y} related to \textit{Society of Jesus}, but since it cannot interpret the entity's semantics, it overlooks the correct triple: (\textit{m.0rgks1y}, \textit{religion.religious\_organization\_leadership.jurisdic} \textit{tion, Argentina}). Additionally, (\textit{m.0rgks1y}, \textit{religion.religious\_organization\_leadership.leader}, \textit{Pope Francis}), (\textit{Pope Francis}, \textit{people.person.birth}, \textit{Italy}), and (\textit{Italy}, \textit{location.country.currency\_used}, \textit{Euro}) lead ChatGPT to assume that Pope Francis, as a leader of the Society of Jesus, is the key to solving the query and conclude that the answer is Euro based on Pope Francis’ birthplace, Italy. This error highlights the semantic gap between the query and the vanilla KG, resulting in incorrect reasoning.
EoG bridges this semantic gap by leveraging structural and feature attributes like similarity, symmetry, transitivity, and hierarchy. Through symmetry, EoG generates the triple: (\textit{m.0rgks1y}, \textit{religion.religious\_organization\_lead-}\newline 
\textit{ership.organization}, \textit{Society of Jesus}), indicating the leadership relationship between \textit{Society of Jesus} and \textit{m.0rgks1y}. Hierarchy generates: (\textit{m.0rgks1y}, \textit{Hypernym\_isA}, \textit{Religious Organization}), clarifying the meaning of \textit{m.0rgks1y}. Transitivity further establishes the relationship: (\textit{Society of Jesus}, \textit{religious\_leadership\_of\_place}, \textit{Argentina}), which aligns semantically with the query, reduces reasoning hops, and enables LLMs to correctly infer that \textit{Society of Jesus} is based in \textit{Argentina}. Finally, the hierarchical triple: (\textit{Argentina}, \textit{Inclusion\_hasContext}, \textit{Argentine Peso}) helps the LLM deduce that the currency of \textit{Argentina} is \textit{Argentine Peso}. This proves that EoG’s query-aligned graph effectively bridges the semantic gap, simplifies reasoning complexity, and enhances LLM efficiency.

In the second example, the vanilla KG contains ambiguous information: (\textit{Eclipse, film.film.starring, m.075wxc2}), and incorrect triples: (\textit{Eclipse, book.book.characters, Leah Clearwater}), (\textit{New Moon, book.book.characters, Leah Clearwater}). These lead ChatGPT to assume that \textit{Leah Clearwater’s} presence in both \textit{Eclipse} and \textit{New Moon} implies a chronological relationship, resulting in the wrong answer: \textit{New Moon}.
In contrast, EoG's query-aligned graph generates: (\textit{Eclipse}, \textit{film.film.prequel}, \textit{The Twilight Saga: New Moon}), which clarifies the prequel relationship between \textit{Eclipse} and the correct answer, \textit{The Twilight Saga: New Moon}. Additionally, the transitive triple: (\textit{Eclipse}, \textit{features\_actor}, \textit{Taylor Lautner}) helps LLMs verify the connection between \textit{Taylor Lautner} and \textit{Eclipse} from the query, leading to the correct reasoning and answer.

These examples demonstrate that EoG-generated query-aligned graphs not only bridge semantic gaps but also reduce reasoning complexity, enabling LLMs to perform more accurate and efficient reasoning.

\section{Impact of different base models and temperature parameters}
\label{plug-and-play-base-model}


\begin{table}[]
\resizebox{\linewidth}{!}{
\begin{tabular}{lcccc}
\hline
\multicolumn{1}{c}{\multirow{2}{*}{Model}} & \multicolumn{2}{c}{CWQ} & \multicolumn{2}{c}{WebQSP} \\ 
\multicolumn{1}{c}{} & \textit{Hit@1} & \textit{F1} & \textit{Hit@1} & \textit{F1} \\ \hline
\multicolumn{5}{c}{\textit{Opensource  LLMs of EoG}} \\ \hline
Qwen-2-7b-instruct & 61.5 & 59.3 & 77.2 & 66.2 \\
LLaMA-3.1-8b-instruct & 70.9 & 64.5 & 81.6 & 69.9 \\ \hline
\multicolumn{5}{c}{\textit{Closed Source LLMs of EoG}} \\ \hline
GPT-4o-mini & 70.8 & 65.1 & 85.0 & 74.1 \\
{GPT-4o-mini}$^*$ & 70.9 & 65.1 & 85.1 & 74.8 \\
GPT-4o-mini\textdagger & 70.6 & 64.8 & 84.8 & 73.8 \\
GPT-3.5-turbo & 65.6 & 60.7 & 83.8 & 73.2 \\
GPT-4o & 72.1 & 66.0 & 82.9 & 71.4 \\
o3 & \textbf{77.6} & \textbf{71.0} & \textbf{85.9} & \textbf{75.1} \\ \hline
\end{tabular}
}
\caption{Impact of different base models and temperature parameters. * indicates temperature is 0.5, while \textdagger \quad indicates temperature is 0.7.}
\label{fig:Impact of different base models and temperature parameters}
\end{table}

As shown in Tab.~\ref{fig:Impact of different base models and temperature parameters}, we supplement our experiments by including the following additional LLM backbones: GPT-3.5-turbo, GPT4o, o3, LLaMA-3.1-8b-instruct, and Qwen2-7b-instruct. For the open-source small models LLaMA-3.1-8b-instruct and Qwen2-7b-instruct, we maintain the same training-free setup. This allows us to conduct a comprehensive evaluation across both open-source and closed-source LLMs, with varying model sizes and inference capabilities. The experimental results show that under our proposed framework, all LLM backbones demonstrate strong performance. Notably, even smaller parameter models like LLaMA and Qwen achieve performance comparable to advanced methods, despite using a training-free setup. Additionally, experiments with temperature settings of 0.2 (default), 0.5, and 0.7 showed differences below 1\%, verifying EoG's reproducibility and plug-and-play flexibility.


\section{Performance on Different KGs}

We supplement our experiments with additional SOTA evaluations. Specifically, we conducted experiments on larger KGQA benchmarks, including GrailQA \cite{10.1145/3442381.3449992} (based on Freebase) and QALD10-en \cite{perevalov2022qald9plusmultilingualdatasetquestion} (on the larger knowledge graph Wikidata \cite{10.1145/2629489}). Combined with the original experimental results in Tab. \ref{main result}, our method demonstrates consistent superiority across KGQA benchmarks of varying scales, further validating its strong generalization ability. Specifically, as shown in Tab. \ref{table:grailqa}, on GrailQA, our method improves Hit@1 scores by 12.5\%, 6.8\%, 16.8\%, and 7.7\% compared to TIARA \cite{shu-etal-2022-tiara}, DECAF \cite{yu2023decafjointdecodinganswers}, ToG, and DoG, respectively, and outperforms TIARA on F1 score by 1.1\%. As shown in Tab. \ref{table:qald10-en}, on the QALD10-en dataset, our method achieves a 12.1\% improvement over ToG.

\begin{table}[]
\centering
\resizebox{0.5\linewidth}{!}{
\begin{tabular}{lcc}
\hline
\multicolumn{1}{c}{\multirow{2}{*}{Model}} & \multicolumn{2}{c}{GrailQA} \\
\multicolumn{1}{c}{} & \textit{Hits@1} & \textit{F1} \\ \hline
\multicolumn{3}{c}{\textit{LLMs Methods}} \\ \hline
IO & 29.4 & - \\
CoT & 28.1 & - \\ \hline
\multicolumn{3}{c}{\textit{LLMs+KG Methods}} \\ \hline
TIARA & 73.0 & 78.5 \\
DECAF & 78.7 & - \\
DoG & 68.7 & - \\
ToG & 77.8 & - \\
EoG & \textbf{85.5} & \textbf{79.6} \\ \hline
\end{tabular}
}
\caption{Performace comparision between advanced methods and EoG on GrailQA. Specifically, we followed ToG and randomly sampled 1000 examples from GrailQA for testing. Best results are in bold.}
\label{table:grailqa}
\end{table}

\begin{table}[]
\centering
\resizebox{0.45\linewidth}{!}{
\begin{tabular}{lc}
\hline
\multicolumn{1}{c}{\multirow{2}{*}{Model}} & QALD10-en \\
\multicolumn{1}{c}{} & \textit{Hits@1} \\ \hline
\multicolumn{2}{c}{\textit{LLMs Methods}} \\ \hline
IO & 42.0 \\
CoT & 62.3 \\ \hline
\multicolumn{2}{c}{\textit{LLMs+KG Methods}} \\ \hline
ToG & 50.2 \\
EoG & \textbf{62.3} \\ \hline
\end{tabular}
}
\caption{Performace comparision between advanced methods and EoG on QALD10-en. Best results are in bold.}
\label{table:qald10-en}
\end{table}

\section{Elaboration of Focus-Aware Multi-Channel Pruning}
\label{elaboration-Focus-Aware Multi-Channel Pruning}
In this section, we will elaborate on the design rationale of Multi-Channel Pruning. Assuming the target answer triple for a query is ($e_1$, $r$, ?), where $e_1$ is the topic entity in the query, we describe three possible situations:
\begin{itemize}
\item \textbf{Situation 0}: The answer triple in the KG is ($e_1$, $r$, $e_2$).
\item \textbf{Situation 1}: The answer triple in the KG is ($e_1$, $r'$, $e_2$), where $r'$ is semantically similar to r but not strongly related.
\item \textbf{Situation 2}: The answer path involves two triples: ($e_1$, $r$, mid) and (mid, $r'$, $e_2$), where mid is an intermediatse entity, and $r'$ is semantically similar to $r$ but not strongly related.
\end{itemize}
Vanilla pruning effectively retrieves ($e_1$, $r$, $e_2$) in Situation 0. However, it often fails to retrieve ($e_1$, $r'$, $e_2$) in Situation 1 and (mid, $r'$, $e_2$) in Situation 2 because these triples are not semantically similar to the query ($e_1$, $r$, ?), resulting in lower similarity rankings. This failure to retrieve the correct triples leads to focus mismatch and often retrieves noise instead of answer triples.
To address these issues, our Three Masking channels are as follows:
\begin{itemize}
\item ($e_1$, $r$, MASK)

\item (MASK, $r$, $e_2$)

\item (MASK, $r$, MASK)
\end{itemize}
The ($e_1$, $r$, MASK) channel and (MASK, $r$, $e_2$) channel focus solely on the head or tail entity and the relation, thereby effectively retrieving triples like ($e_1$, $r'$, $e_2$) in Situation 1. The (MASK, $r$, MASK) channel focuses solely on the relation, thereby effectively retrieving triples like (mid, $r'$, $e_2$) in Situation 2. By combining the strengths of all three channels, the Three Masking mechanism can reliably retrieve the desired answer triples.

To better illustrate the contribution of each channel, we supplemented the paper with channel contribution experiments. We use the MRR (Mean Reciprocal Rank) metric to measure the ability of each channel to rank answer triples (i.e., the essential triples needed to answer the query). A higher MRR indicates that the answer triples are ranked higher within the channel and are more likely to be retrieved. We conducted these experiments on the WebQSP and CWQ datasets, as shown in Tab. \ref{table:three-channel-pruning}. Results show that a single channel's MRR score is similar to vanilla pruning, but combining three channels achieves the highest MRR, exceeding vanilla pruning by 0.002. This demonstrates that the Three Masking Pruning more effectively mitigates the focus mismatch.

\begin{table}[]
\centering
\resizebox{\linewidth}{!}{
\begin{tabular}{lcc}
\hline
\multicolumn{1}{l}{Pruning Method} & WebQSP & CWQ \\ \hline
Vanilla Pruning ($e_1$,$r$,$e_2$) & 0.0058 & 0.0049 \\
Channel ($e_1$,$r$,MASK) & 0.0056 & 0.0048 \\
Channel 2(MASK,$r$,$e_2$) & 0.0055 & 0.0051 \\
Channel 3 (MASK,$r$,MASK) & 0.0071 & 0.0059 \\
Three Channel Pruning (Combined) & 0.0078 & 0.0075 \\ \hline
\end{tabular}
}
\caption{MRR scores comparision between vanilla pruning and our Three Channel Pruning.}
\label{table:three-channel-pruning}
\end{table}

We demonstrate the contribution of each channel by calculating the proportion of its MRR score to the sum of MRR scores across all channels, as shown in Tab. \ref{table:channel-contrbution}. We found that the (MASK, $r$, MASK) channel contributes significantly more on both datasets than the other two channels, proving its effectiveness in addressing Situation 2.

\begin{table}[]
\centering
\resizebox{\linewidth}{!}{
\begin{tabular}{lccc}
\hline
\multicolumn{1}{l}{Dataset} & ($e_1$,$r$,MASK) & (MASK,$r$,$e_2$) & (MASK,$r$,MASK) \\ \hline
WebQSP & 30.72\% & 30.08\% & 39.18\% \\
CWQ & 30.48\% & 32.28\% & 37.22\% \\ \hline
\end{tabular}
}
\caption{Contribution of each channel.}
\label{table:channel-contrbution}
\end{table}

\section{Error Analysis}

We supplement our paper with an error analysis during the enrichment process. Specifically, we categorized errors during enrichment into three types:

\begin{itemize}
    \item \textbf{Identify Errors}: LLMs enrich triples that are unrelated to the query.

    \item \textbf{Structural Enrich Errors}: LLMs introduce incorrect or query-irrelevant relationships/paths during structural enrichment.

    \item \textbf{Feature Enrich Errors}: LLMs generate incorrect or query-irrelevant ontology during feature enrichment.
\end{itemize}

To analyze these errors, we randomly selected 100 samples from the WebQSP dataset and conducted a statistical evaluation, as shown in Tab. \ref{table:error-analysis-enrich}.

\begin{table}[]
\centering
\resizebox{\linewidth}{!}{
\begin{tabular}{ccc}
\hline
identify errors & structural enrich errors & feature enrich errors \\ \hline
8 & 1 & 2 \\ \hline
\end{tabular}
}
\caption{Error analysis during enrichment process.}
\label{table:error-analysis-enrich}
\end{table}

Our results indicate that identify errors are the primary cause of hallucinated knowledge during the enrichment process. This suggests that there is space for improvement in LLMs' ability to accurately identify query-relevant triples.

To demonstrate that our method effectively mitigates the hallucination problem of LLMs in KGQA, we performed an error analysis comparing our method with baseline methods. Following the error taxonomy from ToG, we categorized the errors: \textbf{1.Hallucination}; \textbf{2.Refuse Error}; \textbf{3.Format Error}. On the WebQSP dataset, we randomly selected 100 samples for statistical analysis, as shown in Tab. \ref{table:error-analysis-cmp}. The results show that our method halves hallucinations compared to standard prompts (IO) and chain of thought (CoT) while significantly reducing refuse errors by leveraging query-aligned KGs.

\begin{table}[]
\centering
\resizebox{\linewidth}{!}{
\begin{tabular}{lccc}
\hline
\multicolumn{1}{c}{Model} & halluciation & refuse error & format errors \\ \hline
IO & 23 & 16 & 13 \\
CoT & 20 & 35 & 13 \\
EoG & 11 & 1 & 11 \\ \hline
\end{tabular}
}
\caption{Error analysis comparison with baseline methods.}
\label{table:error-analysis-cmp}
\end{table}

\end{document}